\pdfoutput=1
\documentclass[11pt]{article}
\usepackage[final]{acl}
\usepackage{times}
\usepackage{latexsym}
\usepackage{authblk}
\usepackage[T1]{fontenc}
\usepackage[utf8]{inputenc}
\usepackage{microtype}
\usepackage{inconsolata}
\usepackage{graphicx}

\usepackage{booktabs}      
\usepackage{adjustbox}     
\usepackage{multirow}      
\usepackage{array}         
\usepackage{graphicx}      
\usepackage{makecell}
\usepackage{xcolor} 
\usepackage{amssymb} 
\usepackage{subcaption}

\usepackage{array}
\usepackage{listings}

\usepackage{xspace}

\lstdefinelanguage{SQL}{
    keywords={SELECT, FROM, WHERE, AND, OR, AS, COUNT, JOIN, ON, GROUP BY, WITH, MAX},
    keywordstyle=\bfseries\color{blue},
    morekeywords={[2]{Medal, Format, Tournament, Athlete}},
    keywordstyle={[2]\color{teal}},
    sensitive=true,
    comment=[l]{--},
    morecomment=[s]{/*}{*/},
    commentstyle=\color{gray},
    stringstyle=\color{purple},
    morestring=[b]',
}

\newcommand{\datasetName}{{\textsc{TempTabQA-C}}\xspace}

\title{LLM-Symbolic Integration for Robust Temporal Tabular Reasoning}

\author{
    \textbf{Atharv Kulkarni\textsuperscript{1}\thanks{Work done during internship at UPenn}}, 
    \textbf{Kushagra Dixit\textsuperscript{1}}, \\
    \textbf{Vivek Srikumar}\textsuperscript{1},
    \textbf{Dan Roth}\textsuperscript{2},
    \textbf{Vivek Gupta\textsuperscript{3}\thanks{Primary mentor and Corresponding author.}}\\
    \textsuperscript{1}University of Utah,
    \textsuperscript{2}University of Pennsylvania,
    \textsuperscript{3}Arizona State University\\
    \texttt{\small \{atharvk, danroth\}@seas.upenn.edu} \\
    \texttt{\small kushagra.dixit@utah.edu} \quad \texttt{\small svivek@cs.utah.edu} \\ \texttt{\small vgupt140@asu.edu}}

\begin{document}

\maketitle

\begin{abstract}
Temporal tabular question answering presents a significant challenge for Large Language Models (LLMs), requiring robust reasoning over structured data—a task where traditional prompting methods often fall short. These methods face challenges such as memorization, sensitivity to table size, and reduced performance on complex queries. To overcome these limitations, we introduce \datasetName, a synthetic dataset designed for systematic and controlled evaluations, alongside a symbolic intermediate representation that transforms tables into database schemas. This structured approach allows LLMs to generate and execute SQL queries, enhancing generalization and mitigating biases. By incorporating adaptive few-shot prompting with contextually tailored examples, our method achieves superior robustness, scalability, and performance. Experimental results consistently highlight improvements across key challenges, setting a new benchmark for robust temporal reasoning with LLMs. {Code and \datasetName dataset:} \url{https://coral-lab-asu.github.io/llm_symbolic}.

\end{abstract}

\section{Introduction}

\begin{figure}[htbp]
\vspace{-1.0em}
\begin{center}
    \includegraphics[width=0.60\columnwidth]{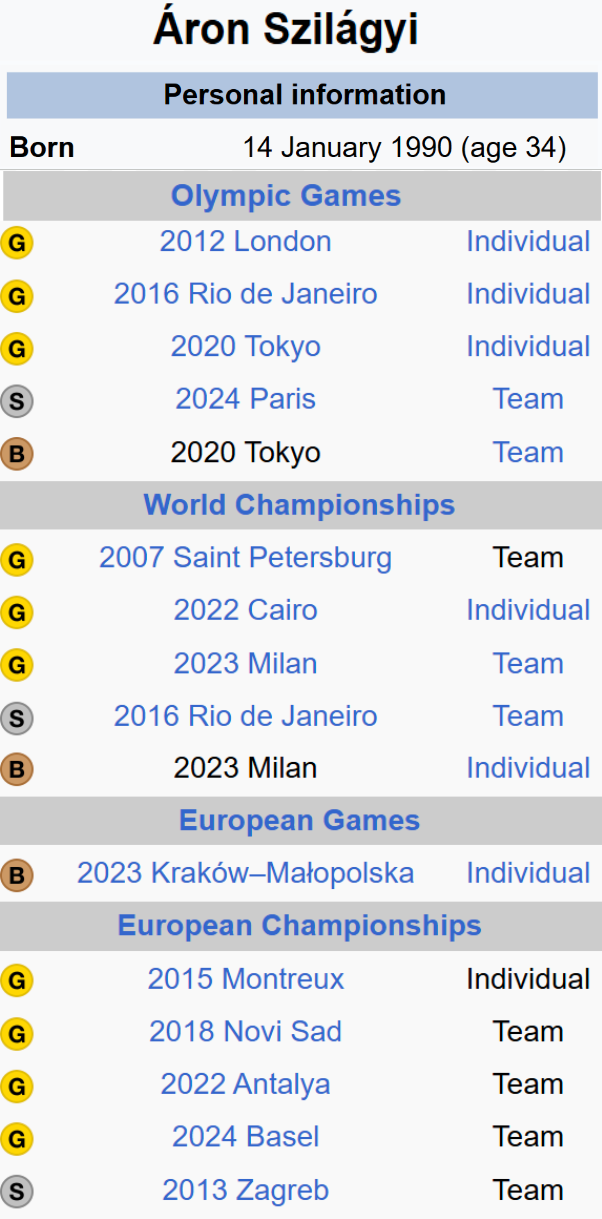} 
\vspace{0.25em}

\noindent \textit{\small \textbf{Q.} In which year did Áron Szilágyi achieve his personal highest number of gold medal wins? \textbf{Gold Answer.} 2022}
\end{center}
\vspace{-0.15em}
\raggedright
\small{
\textbf{\small Direct prompt (CoT):} {\scriptsize In 2020, Áron Szilágyi won a Gold Medal in the Individual format at the Tokyo Olympics. This was his most successful year in terms of highest Gold Medal wins since he did not win multiple Gold Medals in any other format or competition.} \textbf{Answer:} 2020.}\\
\vspace{0.15em}
\textbf{Symbolic Intermediate Representation (SQL):} \\
\vspace{-0.25em}
\begin{lstlisting}[language=SQL]
WITH gold_medal_counts AS (
    SELECT m.year, COUNT(m.medal_id) AS gold_medals
    FROM Medal m JOIN Format f ON m.format_id = f.format_id
    JOIN Tournament t ON f.tournament_id = t.tournament_id
    JOIN Athlete a ON t.athlete_id = a.athlete_id
    WHERE a.name = 'Áron Szilágyi' AND m.type = 'MedalGold'
    GROUP BY m.year )
SELECT year FROM gold_medal_counts
WHERE gold_medals = (SELECT MAX(gold_medals) FROM gold_medal_counts);
\end{lstlisting}
\vspace{-0.25em}
\textbf{Answer:} 2022
\label{fig:aron_table}
\caption{ \small \small Structured table of Áron Szilágyi's achievements with question and answers. Direct prompting fails, whereas, Symbolic Intermediate Representation give correct answer.}
\label{fig:example_figure}
\end{figure}

Large Language Models (LLMs) have revolutionized natural language processing, achieving remarkable success across a variety of tasks. However, answering questions about temporal tabular data task that requires precise reasoning over structured information with time-based attributes remains a significant challenge. This capability is crucial in fields such as finance, healthcare, and policymaking, where actionable insights often depend on understanding and analyzing evolving datasets. Yet, existing methods often fall short, struggling with complex queries, large datasets, and scenarios that require nuanced reasoning. An example of such task, with a long table and corresponding query and it's answer is shown in Figure \ref{fig:example_figure}.

These limitations underscore the need for robust, scalable, and interpretable solutions. A key obstacle lies in the lack of benchmarks that adequately capture the complexity and diversity of temporal reasoning tasks. Existing benchmarks, typically created manually, are inconsistent and fail to provide the variability needed to thoroughly evaluate models. Without rigorous evaluation frameworks, it becomes difficult to diagnose weaknesses or ensure models can handle real-world scenarios. This raises an essential question: \textit{How can we design benchmarks that effectively evaluate temporal reasoning across a range of challenging contexts?}

Equally important is the need for robust methods. Many existing approaches rely on direct prompting, which often depends on heuristics and memorized patterns rather than true reasoning. This results in semantic biases and poor performance in demanding scenarios, such as counterfactual reasoning, large table contexts, or multi-step queries. This leads to a second critical question: \textit{How can we develop methods that remain robust across diverse table structures, dynamic data, and complex queries?}

To address these challenges, we propose a comprehensive framework that reimagines how LLMs approach temporal tabular data. At its core is \textbf{\datasetName}, a synthetic dataset generation method designed to fill the gaps in existing benchmarks. \datasetName provides precise control over data characteristics, enabling consistent and systematic evaluation across a wide range of scenarios, including counterfactual reasoning and intricate temporal queries. Building on this foundation, we introduce a \textbf{symbolic intermediate representation approach} that transforms unstructured tables into structured database schemas. LLMs are guided to generate SQL queries based on these schemas, which are executed to produce accurate answers. E.g. in Figure \ref{fig:example_figure}, the SQL query serves as a symbolic representation and provides the correct answer, whereas direct prompting fails on given query. This structured pipeline reduces semantic biases, enhances interpretability, and significantly improves the generalization of models across different table configurations. Additionally, we incorporate \textbf{adaptive few-shot prompting}, a dynamic approach that selects contextually relevant examples tailored to each query. This method overcomes the limitations of static examples, further improving the robustness of the system in complex scenarios.

Our experiments demonstrate that this framework delivers substantial improvements over direct prompting methods. It excels in critical areas such as counterfactual reasoning, scalability to larger datasets, and the handling of complex queries. Beyond these technical advancements, our work establishes a new benchmark for temporal tabular question answering by addressing fundamental weaknesses in existing approaches and introducing innovative tools for evaluation and reasoning. These contributions pave the way for building more interpretable, scalable, and robust AI systems with  implications for  critical real-world applications. Our contributions are as follows:

\begin{enumerate} 
\vspace{-0.5em}
\setlength\itemsep{0.25em}

\item We introduce \textbf{\datasetName}, a synthetic dataset designed for \textbf{precise and robust evaluation} of temporal tabular reasoning across diverse and challenging scenarios. \item We analyze the \textbf{limitations of direct prompting}, including reliance on \textbf{memorization}, sensitivity to \textbf{table size}, and struggles with \textbf{complex multi-step or counterfactual} reasoning. \item We propose a \textbf{symbolic intermediate representation approach} that enhances interpretability, reduces biases, and improves generalization by guiding LLMs to generate and execute SQL queries on structured schemas. \item We enhance this approach with \textbf{adaptive few-shot prompting}, enabling context-specific example selection for improved flexibility and performance in diverse scenarios. 
\vspace{-0.25em}
\end{enumerate}

\section{The \datasetName Dataset}

The \datasetName dataset is a large-scale, semi-automatically generated resource designed for evaluating temporal reasoning in Large Language Models (LLMs). It provides a benchmark for analyzing the tabular temporal abilities of LLMs by enabling controlled variations in data characteristics, making it superior to traditional human-curated datasets. This section describes the dataset creation process, its schema, and key characteristics.

\subsection{\datasetName Creation Pipeline}

The creation of \datasetName follows a systematic pipeline to extract, structure, and store temporal information from Wikipedia infoboxes. Below, we describe the steps involved in detail.

\paragraph{Extracting Temporal Information.}

Temporal information about athletes, tournaments, events, and achievements is extracted from Wikipedia infoboxes. These tables contain attributes such as \textit{"Name," "Date of Birth," "Tournaments Played,"} and \textit{"Medals Won,"} which are programmatically extracted and input into a relational database using a predefined schema. This step ensures that the raw tabular data is converted into a structured format for efficient querying and storage.

\paragraph{Relational Database Creation.}

The structured temporal data is converted into a relational database schema to enable efficient storage and querying. The schema is designed to represent key entities and their relationships comprehensively:

\begin{itemize}
\vspace{-0.5em}
\setlength\itemsep{0.25em}
    \item \textbf{Athlete Table:} Contains a unique \texttt{athlete\_id} and the corresponding athlete's name.
    \item \textbf{Personal Information Table:} Captures birth year, month, and day for each athlete, linked to the \texttt{athlete\_id}.
    \item \textbf{Tournament Table:} Stores tournament details, such as the name (e.g., "Olympic Games") and the \texttt{athlete\_id}.
    \item \textbf{Format Table:} Represents event formats (e.g., "100m Freestyle"), linked to tournaments through \texttt{tournament\_id}.
    \item \textbf{Medal Table:} Documents medals, including type (e.g., "Gold"), year (e.g., "2016"), and location (e.g., "Rio de Janeiro"), linked to formats through \texttt{format\_id}.

\vspace{-0.5em}
\end{itemize}

This schema ensures all entities are interconnected via primary and foreign keys, enabling complex queries like calculating an athlete’s age at the time of their first medal or comparing performance across tournaments.

\paragraph{Question and Answer Generation}

Questions are generated using predefined templates filled with key attributes from the relational database. Templates capture a wide range of temporal reasoning scenarios, such as:

{ 
\footnotesize
\begin{itemize}
\vspace{-0.5em}
\setlength\itemsep{-0.25em}
    \item At what age did [\textcolor{blue}{Athlete}] win his most recent [\textcolor{red}{Tournament}] [\textcolor{green!40!black!60}{Medal Type}]?
    \item At what age did \textcolor{blue}{Michael Phelps} win his most recent \textcolor{red}{Pan Pacific Championships} \textcolor{green!40!black!60}{Silver Medal}?
\vspace{-0.5em}
\end{itemize}
}

To generate answers, the relational database is queried using SQL-based logic, which systematically retrieves the necessary information. For instance, answering a question about the age of an athlete during a specific tournament involves retrieving the athlete’s birth year and the tournament year from the database and calculating the difference. Similarly, questions about medal counts or locations are answered by aggregating or filtering data from the tables.

The SQL-based logic is generalized across various question types, allowing the generation of thousands of unique question-answer pairs. Examples include:

{ 
\vspace{-0.5em}
\footnotesize
\begin{itemize}
\setlength\itemsep{-0.25em}
    \item At what age did \textcolor{blue}{Michael Phelps} win his most recent \textcolor{red}{Olympic Games} \textcolor{green!40!black!60}{Silver Medal}? Answer: \textcolor{purple}{29}
    \item In which city did \textcolor{blue}{Caeleb Dressel} win his most recent \textcolor{red}{Olympic Games} \textcolor{green!40!black!60}{Silver Medal}? Answer: \textcolor{purple}{Tokyo}
\vspace{-0.5em}
\end{itemize}
}
This approach ensures the dataset is both scalable and robust for evaluating temporal reasoning in LLMs.

\subsection{\datasetName Composition and Splits}

The \datasetName dataset is divided into \textbf{Original} and \textbf{CounterFact} questions, with each category further subdivided based on table size and question reasoning difficulty. This structure enables fine-grained and comprehensive evaluations.

\paragraph{Original Questions.}  
Original questions are derived directly from the structured database and are categorized as follows:

\vspace{-0.5em}
\subparagraph{1. Table size:} we make questions on the table with varied sizes: (a.) \textbf{Small Tables:} Contain concise data, typically representing athletes with fewer medals, (b.) \textbf{Large Tables:} Contain extensive data, often representing athletes with a larger number of medals.

\vspace{-0.5em}
\subparagraph{2. Question Complexity:} we answer questions on varied difficulty some requiring complex multi-hop reasoning: (a.) \textbf{Easy:} Require basic facts retrieval or single-step reasoning. E.g.: \emph{"How many formats has Michael Phelps played?"}, (b.) \textbf{Medium:} Involve multi-step reasoning, such as calculations or comparisons. E.g.: \emph{"At what age did Michael Phelps win his most recent Olympic Silver Medal?"}, and (c.) \textbf{Hard:} Demand complex reasoning, temporal analysis, and synthesis of multiple facts.  E.g.: \emph{"What is the shortest time span (in years) within which Michael Phelps won gold, silver, and bronze medals in the same format across any tournament?"}

\paragraph{Counterfactual Questions.}  
Counterfactual questions modify specific facts in the original dataset while maintaining the same categorization based on table size and difficulty of the reasoning of the questions. This design challenges models to reason effectively under hypothetical scenarios.

\paragraph{Significance of \datasetName.}  
The dataset offers several unique advantages: (a.) \textbf{Controlled Evaluation:} Provides a framework for systematically testing LLMs across diverse data characteristics., (b.) \textbf{Scalability:} Comprises over 200,000 questions spanning a wide range of contexts and complexities., and (c.) \textbf{Fine-Grained Analysis:} Facilitates benchmarking of model biases and limitations, particularly for temporal reasoning. 

By providing a controlled, scalable, and diverse dataset, \datasetName establishes a robust foundation for advancing research on temporal reasoning in LLMs.

\subsection{\datasetName Test Statistics.} In order to evaluate the LLM's we created a subset of the \datasetName dataset having the following number of question per category:

\begin{table}[htbp]
\small
\centering
\begin{tabular}{lc|lc}
\toprule
\textbf{Category} & \textbf{\#Examples} & \textbf{Category} & \textbf{\#Examples}\\
\midrule
Original & 578 & Easy & 732 \\
Counterfactual & 699 & Medium & 507 \\
Small Table & 855 & Hard & 719 \\
Large Table & 538 & \textbf{Total} & \textbf{5067} \\
\bottomrule
\end{tabular}
\caption{ \small Dataset Splits and Their Number of Examples.}
\label{tab:dataset_splits}
\vspace{-0.5em}
\end{table}

In our test set, the average context length for the \textbf{Small Table Split} is \textbf{53.80} words when using the infobox as the context, while for the \textbf{Large Table Split}, it increases significantly to \textbf{348.85 words.}

\section{Answering Questions about Tables}
\label{sec:answering_tables}

In this section, we describe two approaches for answering questions about structured data.

\subsection{Approach 1: Direct Prompting}
Direct prompting is the standard approach today, where the LLM is presented with the table contents alongside a natural language question and is expected to return an answer. To improve reasoning, prompting methods have been proposed, such as Chain of Thought (CoT), Program of Thought (PoT), Faithful Chain of Thought (F-CoT), and Plan and Solve, etc.

While these methods introduce symbolic elements and explicit reasoning, they still rely on raw table data and suffer from limitations. The model may rely on memorized patterns rather than reasoning about the content, and be sensitive to table size and inconsequential perturbations in the table. Moreover, although CoT, PoT and F-CoT provide explicit reasoning, the reasoning remains loosely structured, making verification and consistency challenging.

Despite these issues, direct prompting remains widely used due to its ease of implementation.

\subsection{Approach 2: Symbolic Intermediate Representation}
We propose an alternative approach: \textbf{symbolic intermediate representation}. We hypothesize that it can mitigate the issues of direct prompting by shifting from raw table data processing to structured query generation.

Instead of providing raw table contents, we transform the table into a \textbf{structured schema}, which consists of metadata such as column names, data types, and relational links, without exposing actual values. The LLM then builds its reasoning upon this schema by generating a \textbf{structured query (e.g., SQL) to retrieve the answer.}

By passing only the schema, the LLM is guided to reason in a structured manner, reducing reliance on spurious table patterns. The model’s generated query is executed on the database to retrieve the final answer, ensuring explicit and verifiable reasoning.

With this approach, since the raw data is masked, the model is less likely to hallucinate due to table noise, and less sensitive to minor table modifications. Moreover, the time to answer a question depends less on the size of the table, because query execution in database engines is efficient even for large tables. Finally, the generated query provides a highly structured reasoning path, ensuring systematic verification and consistency.

\section{Experimental Setup}
We designed experiments to address the following research questions:

\begin{enumerate}
    \item \textbf{Robustness to Counterfactual Data}: How robust are direct LLM prompts to counterfactual data, and can symbolic intermediate representations improve this?
    \vspace{-0.25em}
    \item \textbf{Handling Large Tables}: Can a symbolic intermediate representation outperform direct prompting when applied to larger tables?
    \vspace{-0.25em}
    \item \textbf{Impact of Question Complexity}: How does increasing question complexity impact the performance of these two approaches?
\end{enumerate}

To answer these questions, we conducted experiments to evaluate the two core approaches: \textbf{Direct Prompting} and \textbf{Symbolic Intermediate Representation}.

\subsection{Direct Prompting}
In this approach, models are provided with the table and question in natural language, and they generate answers as free text or structured programs based on the raw table. We evaluate multiple configurations. In the \textbf{static few-shot} setup, the model is presented with a fixed set of examples, whereas in the \textbf{adaptive few-shot} setup, examples are dynamically selected based on their relevance to the given question.

Beyond few-shot prompting, we evaluate \textbf{C.L.E.A.R}, a structured method that first extracts relevant rows from the table, decomposes the question into sub-questions, solves each sub-question individually, and synthesizes the final answer. We also consider \textbf{Program of Thought (PoT)}, where the model generates structured Python programs to extract the necessary table contents and store them as variables before computing the answer. \textbf{Faithful Chain of Thought (FCoT)} extends PoT by requiring the model to decompose the reasoning process into explicit steps before generating the corresponding program. We also evaluate \textbf{Chain of Thought (CoT)}, where the model explicitly generates intermediate logical steps before producing the final answer, and \textbf{Plan and Solve}, a two-stage approach where the model first formulates a reasoning plan before executing step-by-step calculations.

\subsection{Symbolic Intermediate Representation}
In contrast to direct prompting, this approach does not expose the raw table contents to the model. Instead, the model is provided with only the table schema and must generate an SQL query, which is executed to retrieve the answer. We evaluate two variations: \textbf{static few-shot SQL}, where a fixed set of natural language-to-SQL mappings is included in the prompt alongside the schema, and \textbf{adaptive few-shot SQL}, where SQL examples are dynamically selected based on their relevance to the given question. In both cases, the model receives the schema as context and must generate an appropriate SQL query to obtain the answer.

We analyze the performance of these methods in the \textit{Results and Analysis} section.

We used the \datasetName dataset, which includes Original, counterfactual, and question difficulty (Easy, Medium, Hard) splits, along with small and large table contexts. Models were evaluated using Exact Match Score (EMS) \footnote{We also used a relaxed version of EMS (REMS), with similar results detailed in the appendix.}, focusing on the following key splits:

\begin{itemize}
\vspace{-0.5em}
\setlength\itemsep{-0.25em}
    \item \textbf{Original vs. Counterfactual}: We examined whether the gap between Original and Counterfactual data reduces as we move toward symbolic intermediate reasoning.
    \item \textbf{Large Table vs. Small Table}: We evaluated if the gap between large and small tables decreases with symbolic intermediate reasoning.
    \item \textbf{Performance by Question Complexity}: We analyzed performance trends across Easy, Medium, and Hard questions, particularly the improvement brought by symbolic intermediate reasoning.
\end{itemize}

Through these experiments, we aim to demonstrate that symbolic intermediate reasoning reduces sensitivity to counterfactual data, scales better with table size, and handles increasing question complexity more effectively than direct prompting.

\section{Results and Analysis}

In this section, we present the results for GPT-4o and Gemini 1.5 Pro. Additionally, we evaluated Gemini 1.5 Flash, GPT-4o Mini, Mixtral, Llama 3.1 70B, Code Llama, and SQL Coder, which demonstrated similar trends. The results of these additional experiments are included in the Appendix.

\subsection{Robustness on Counterfactual Data.}

To evaluate counterfactual robustness, we compare model performance on original and counterfactual datasets. Table \ref{tab:gpt_4o_original_counterfactual_gap} and \ref{tab:gemini_1_5_pro_original_counterfactual_gap} summarizes these results, including the performance gap ($\Delta$) between the original and counterfactual performance for GPT 4o and Gemini-1.5-Pro.

\renewcommand{\arraystretch}{1.0}
\begin{table}[htbp]
\centering
\footnotesize
\begin{tabular}{l|l|c|c|c}
\toprule
\scriptsize{\textbf{Approach}} & \scriptsize{\textbf{Method}} & \scriptsize{\textbf{Original}} & \scriptsize{\textbf{CounterF.}}  & \scriptsize{\textbf{$\Delta$}} \\
\midrule
\multirow{7}{*}{Direct} & Static & 56.57 & 42.35 & 14.22 \\
 & Adaptive & 58.13 & 40.92 & 17.21 \\
 & C.L.E.A.R & 65.57 & 48.21 & 17.36 \\
 & CoT & 69.90 & 48.50 & 21.40 \\
 & Plan \textbf{$\&$} Solve & 69.55 & 46.50 & 23.05 \\
 & PoT & 56.40 & 47.07 & 9.33 \\
 & Faithful CoT & 57.44 & 47.78 & 9.66 \\
\midrule
\multirow{2}{*}{\bf SQL} & \textbf{Static} & \textbf{65.22} & \textbf{60.94} & \textbf{4.28} \\
 & \textbf{Adaptive} & \textbf{71.63} & \textbf{68.67} & \textbf{2.96} \\
\bottomrule
\end{tabular}
\caption{\footnotesize Original vs Counterfactual for GPT-4o.}
\label{tab:gpt_4o_original_counterfactual_gap}
\vspace{-1.0em}
\end{table}

\renewcommand{\arraystretch}{1.0}
\begin{table}[htbp]
\centering
\footnotesize
\begin{tabular}{l|l|c|c|c}
\toprule
\scriptsize{\textbf{Approach}} & \scriptsize{\textbf{Method}} & \scriptsize{\textbf{Original}} & \scriptsize{\textbf{CounterF.}}  & \scriptsize{\textbf{$\Delta$}} \\
\midrule
\multirow{7}{*}{Direct} & Static & 52.91 & 39.92 & 12.99 \\
 & Adaptive & 53.48 & 44.19 & 9.29 \\
 & C.L.E.A.R & 49.33 & 40.66 & 8.67 \\
 & CoT & 66.46 & 55.75 & 10.71 \\
 & Plan \textbf{$\&$} Solve & 60.75 & 54.31 & 6.44 \\
 & PoT & 53.83 & 46.20 & 7.63 \\
 & Faithful CoT & 53.95 & 47.64 & 6.31 \\
\midrule
\multirow{2}{*}{\bf SQL}& \textbf{Static} & \textbf{59.08} & \textbf{55.52} & \textbf{3.56} \\
 & \textbf{Adaptive} & \textbf{65.29} & \textbf{65.13} & \textbf{0.16} \\
\bottomrule
\end{tabular}
\caption{\footnotesize Original vs Counterfactual for Gemini 1.5 Pro.}
\label{tab:gemini_1_5_pro_original_counterfactual_gap}
\vspace{-1.0em}
\end{table}

\paragraph{Analysis:} Comparing the performance of GPT-4o and Gemini 1.5 Pro across original and counterfactual datasets provides valuable insights into the robustness of Direct Prompting and SQL-based reasoning methods. \textit{A model that truly reasons about data should not be affected by the origin of the data.}

However, for Direct Prompting, both models exhibit significant performance gaps between the original and counterfactual datasets, indicating a heavy reliance on memorized knowledge rather than robust reasoning capabilities. For GPT-4o, the performance gaps are 14.22 (non-adaptive) and 17.21 (adaptive), while Gemini 1.5 Pro shows similar gaps of 12.99 and 9.29, respectively. Notably, the adaptive approach improves the performance on original data but increases sensitivity to counterfactuals for GPT-4o. In contrast, Gemini 1.5 Pro’s adaptive Direct Prompting reduces the gap but still fails to address the core issue of data sensitivity.

On the other hand, SQL-based methods, demonstrate superior robustness in both models, with performance gaps significantly smaller than those in Direct Prompting. For GPT-4o, the non-adaptive SQL gap is just 4.28, and the adaptive SQL gap is 2.96. Similarly, for Gemini 1.5 Pro, the non-adaptive SQL gap is 3.56, and the adaptive SQL gap reduces further to 0.16, showcasing its capability to reason effectively across the original and counterfactual datasets. The use of symbolic intermediate representations in SQL methods explains this robustness, as these approaches operate independently of the data origin, focusing instead on schema-driven reasoning.

Finally, the adaptive approach enhances performance across both methods and models, particularly for SQL. For example, in GPT-4o, adaptive SQL improves counterfactual performance by 7.73 points compared to non-adaptive SQL, while in Gemini 1.5 Pro, it further narrows the performance gap to an almost negligible 0.16 points, while improving counterfactual performance by 9.61 points. This highlights the role of adaptive few-shot examples in enhancing model reasoning capabilities and robustness across diverse datasets.

\subsection{Impact of Table Size.}

To evaluate the impact of data size, we compare model performance on small and large datasets. Table \ref{tab:gpt4o_small_large_gap} and \ref{tab:gemini_1_5_pro_small_large_gap} presents the results, including the gap between small and large datasets for GPT 4o and Gemini 1.5 Pro.

\renewcommand{\arraystretch}{1.0}
\begin{table}[htbp]
\centering
\footnotesize
\begin{tabular}{l|l|c|c|c}
\toprule
\textbf{Approach} & \textbf{Method} & \textbf{Small} & \textbf{Large} & \textbf{$\Delta$} \\
\midrule
\multirow{7}{*}{ Direct} & Static & 71.11 & 46.84 & 24.27 \\
 & Adaptive & 73.92 & 48.88 & 25.04 \\
 & C.L.E.A.R & 76.49 & 53.53 & 22.96 \\
 & CoT & 74.15 & 56.13 & 18.02 \\
 & Plan \textbf{$\&$} Solve & 75.44 & 55.20 & 20.24 \\
 & PoT & 62.22 & 50.19 & 12.03 \\
 & Faithful CoT & 62.69 & 50.19 & 12.50 \\
\midrule
\multirow{2}{*}{\bf SQL} & \textbf{Static} & \textbf{73.57} & \textbf{70.63} & \textbf{2.94} \\
 & \textbf{Adaptive} & \textbf{73.92} & \textbf{72.86} & \textbf{1.06} \\
\bottomrule
\end{tabular}
\caption{\footnotesize Small Table vs Large Table for GPT-4o}
\label{tab:gpt4o_small_large_gap}
\vspace{-1.0em}
\end{table}

\renewcommand{\arraystretch}{1.0}
\begin{table}[htbp]
\centering
\footnotesize
\begin{tabular}{l|l|c|c|c}
\toprule
\textbf{Approach} & \textbf{Method } & \textbf{Small} & \textbf{Large} & \textbf{$\Delta$} \\
\midrule
\multirow{7}{*}{Direct} & Static & 65.02 & 43.86 & 21.16 \\
 & Adaptive & 67.27 & 41.86 & 25.41 \\
 & C.L.E.A.R & 56.85 & 40.95 & 15.90 \\
 & CoT & 77.55 & 57.67 & 19.88 \\
 & Plan \textbf{$\&$} Solve & 73.43 & 56.72 & 16.71 \\
 & PoT & 64.68 & 49.84 & 14.83 \\
 & Faithful CoT & 63.85 & 48.47 & 15.39 \\
\midrule
\multirow{2}{*}{\textbf{SQL}} & \textbf{Static} & \textbf{77.41} & \textbf{71.32} & \textbf{6.09} \\
 & \textbf{Adaptive} & \textbf{75.31} & \textbf{72.43} & \textbf{2.88} \\
\bottomrule
\end{tabular}
\caption{\footnotesize   Small Table vs Large Table for Gemini 1.5 Pro}
\label{tab:gemini_1_5_pro_small_large_gap}
\vspace{-1.0em}
\end{table}

\paragraph{Analysis:}
\textit{A model capable of genuine reasoning should operate independently of data size. For example, the correctness of an SQL query's result is unaffected by the size of the tables—it impacts only the computation time, not the quality of the outcome.} However, the trends for small vs. large tables closely mirror those observed in the original vs. counterfactual analysis. Direct Prompting shows significant performance drops with larger tables, with GPT-4o and Gemini 1.5 Pro exhibiting gaps of 24.27 and 21.16 in non-adaptive settings, respectively. This underscores the method's sensitivity to data complexity and dependence on memory.

In contrast, SQL-based methods demonstrate remarkable robustness, maintaining minimal performance gaps across table sizes (e.g., 1.06 for adaptive SQL in GPT-4o and 2.88 in Gemini 1.5 Pro). This resilience stems from schema-driven reasoning, which abstracts away from the data’s size or origin \footnote{We tested counterfactual versions, showing similar findings to section 4.1.}. Adaptive few-shot examples further enhance performance, particularly for SQL-based methods, allowing them to consistently deliver high accuracy even with larger tables.

These findings emphasize that Direct Prompting struggles with data complexity and scale, mirroring its limitations in counterfactual settings. SQL-based methods, on the other hand, exemplify robustness and scalability by leveraging schema-driven symbolic representations that are agnostic to data size or source. The dynamic selection of adaptive examples further strengthens their reliability, making them a superior choice for reasoning over complex and evolving datasets.

\subsection{Effect of question complexity.}

To evaluate question complexity effects, we compare model performance on Easy, Medium, and Hard questions. Table \ref{tab:gpt4o_easy_medium_hard} and \ref{tab:gemini_1_5_pro_easy_medium_hard} summarizes the results for GPT-4o and Gemini-1.5-Pro respectively.

\renewcommand{\arraystretch}{1.0}
\begin{table}[htbp]
\centering
\footnotesize
\begin{tabular}{l|l|c|c|c}
\toprule
\textbf{Approach} & \textbf{Method} & \textbf{Easy} & \textbf{Medium} & \textbf{Hard} \\
\midrule
\multirow{7}{*}{Direct} & Static & 71.18 & 63.12 & 53.35 \\
 & Adaptive & 74.38 & 63.91 & 54.17 \\
 & C.L.E.A.R & 76.40 & 71.99 & 62.62 \\
 & CoT & 78.43 & 75.35 & 64.06 \\
 & Plan \textbf{$\&$} Solve & 77.96 & 70.81 & 60.25 \\
 & PoT & 69.94 & 57.20 & 48.81 \\
 & Faithful CoT & 69.24 & 56.02 & 47.17 \\
\midrule
\multirow{2}{*}{\textbf{SQL}} & \textbf{Static} & \textbf{78.89} & \textbf{75.15} & \textbf{62.31} \\
 & \textbf{Adaptive} & \textbf{80.06} & \textbf{73.37} & \textbf{66.74} \\
\bottomrule
\end{tabular}
\caption{\footnotesize Easy, Medium, and Hard results for GPT-4o}
\label{tab:gpt4o_easy_medium_hard}
\vspace{-1.0em}
\end{table}

\renewcommand{\arraystretch}{1.0}
\begin{table}[htbp]
\centering
\footnotesize
\begin{tabular}{l|l|c|c|c}
\toprule
\textbf{Approach} & \textbf{Method} & \textbf{Easy} & \textbf{Medium} & \textbf{Hard} \\
\midrule
\multirow{7}{*}{Direct} & Static & 65.79 & 58.53 & 50.00 \\
 & Adaptive & 66.26 & 56.47 & 46.74 \\
 & C.L.E.A.R & 58.38 & 59.77 & 51.14 \\
 & CoT & 83.29 & 72.69 & 65.87 \\
 & Plan \textbf{$\&$} Solve & 79.42 & 72.25 & 63.60 \\
 & PoT & 73.97 & 57.12 & 47.57 \\
 & Faithful CoT & 74.52 & 58.15 & 46.85 \\
\midrule
\multirow{2}{*}{\textbf{SQL}} & \textbf{Static} & \textbf{80.86} & \textbf{70.33} & \textbf{59.59} \\
 & \textbf{Adaptive} & \textbf{75.86} & \textbf{71.47} & \textbf{59.24} \\
\bottomrule
\end{tabular}
\caption{\footnotesize Easy, Medium, and Hard results on Gemini 1.5 Pro}
\label{tab:gemini_1_5_pro_easy_medium_hard}
\vspace{-1.0em}
\end{table}

\paragraph{Analysis:} 
Performance consistently declines across all models and settings as question complexity increases from Easy to Hard, aligning with previous findings on the influence of data size and complexity. \textit{While such a drop is expected for both models and humans (though less severe for the latter), the key question is: can we do better and reduce this decline?} Direct Prompting struggles as question complexity increases, with significant drops in accuracy (e.g., from 71.18 to 53.35 for non-adaptive GPT-4o). Direct adaptive prompting also struggles to mitigate this decline and remains limited in handling complex queries effectively.

SQL-based methods demonstrate greater resilience to complexity, maintaining higher accuracy across all levels. For example, non-adaptive SQL in GPT-4o drops moderately from 78.89 (Easy) to 62.31 (Hard), while adaptive SQL, achieves the best performance of 66.74 on the Hard data split. Similarly, Gemini 1.5 Pro exhibits stable performance with SQL based reasoning.

These results reinforce SQL's robustness through schema-driven reasoning, which abstracts complexity and reduces reliance on memorization. Introducing adaptive examples lifts accuracy across prompting strategies, with the most pronounced gains in the SQL-based variants, which remain the strongest on the hardest queries. This underscores the importance of structured reasoning and adaptive techniques for tackling increasing data and query complexity effectively.

\section{What Did We Learn?}
\paragraph{1. Impact of Symbolic Representations.}
Parsing data into symbolic queries consistently boosts model performance. Symbolic representations bridge counterfactual gaps, reduce dependence on data size, and enhance the handling of complex questions. By structuring data more clearly, symbolic queries improve robustness and address challenges like noise and memorization.

\paragraph{2. Benefits of Schema-Based Reasoning.}
Schemas provide a clean, data-agnostic abstraction of database structures, removing irrelevant noise and simplifying reasoning. By presenting only the schema without any underlying data, we ensure there is no room for memorization. Unlike raw tables, which mix useful and irrelevant data, schemas provide a stable framework that ensures consistent performance, especially in counterfactual scenarios where structured reasoning is critical.

\paragraph{3. Effect of Data Size.}
Data size significantly affects model performance. Larger tables often introduce noise, increasing the risk of hallucinations. Schemas mitigate this by segmenting data into key components, reducing cognitive overload and clarifying the reasoning process, allowing models to perform more reliably on large, complex datasets.

\paragraph{4. Handling Complex Questions.}
Schema-based reasoning excels in answering complex questions by supporting logical, step-by-step reasoning. SQL query generation fosters clarity and reduces ambiguity. In contrast, raw text tables, especially those with counterfactual data, often lack structure, leading to errors or incomplete reasoning. By offering a predefined framework, schemas reduce cognitive demands, enabling models to handle nuanced queries more effectively.

\section{Discussion of Model Failures}

\subsection{Inadequacy of Direct Complex Strategies}

Several techniques, such as Program of Thought (PoT) \citet{chen2023program}, Chain of Table\cite{wang2024chain}, Binder\cite{cheng2023bindinglanguagemodelssymbolic}, Dater \citet{10.1145/3539618.3591708}, and Plan and Solve \citet{wang2023plan}, aim to handle complex queries. However, these methods fall short when detailed query plans are needed. The complexity of tasks involving multiple steps, conditional logic, and dependencies cannot be captured by direct prompting alone. Each query introduces unique variables, making strategies like PoT fails for complex reasoning.

For example, a query requiring the join of three large tables with specific conditions cannot be effectively handled by PoT, which may only generate simple steps like \textit{"select from Table A"} or \textit{"filter Table B."} These methods fail to capture the necessary logic for combining tables or handling multiple joins and nested queries. Such complexity requires a carefully constructed query plan, which direct prompting cannot produce.

The core issue is the complexity of the underlying query plans. PoT may generate query plans, but they struggle with complex operations like joins, aggregations, and nested subqueries, which demand precise sequencing and optimization. Research, particularly by \citet{akioyamen2024unreasonable}, argues that query planning requires structured approaches like SQL to manage these complexities, reinforcing that simpler prompting strategies are insufficient for intricate query reasoning.

\subsection{Challenges with Symbolic Approach}
Despite advancements in symbolic representation, several challenges remain in improving model reliability and performance:

\vspace{-0.3em}
\paragraph{1. SQL Query Inconsistencies}
The model often misuses SQL constructs, such as over-relying on \texttt{LIMIT 1} when multiple answers are needed \textit{"List all the formats in which Carolina Marín has won medals?"} or adding redundant joins that slow execution \textit{"How many tournaments did Michael Phelps win in 2008?"}. It also misaligns query objectives, failing to handle aggregates or \texttt{GROUP BY} clauses properly \textit{"What are the medal counts for each athlete in the Olympics?"}.

\vspace{-0.3em}
\paragraph{2. Temporal and Positional Reasoning Errors}
The model struggles with temporal and positional reasoning, often hallucinating columns or misinterpreting data \textit{"At what age did Michael Phelps win his most recent Olympic Gold Medal?"}. It also misaligns aggregations over time \textit{"Which athlete had the most consistent medal wins over the last decade?"} and hierarchical relationships \textit{"Which was Michael Phelps' most recent tournament medal?"}.

\vspace{-0.3em}
\paragraph{3. Nested and Conditional Logic Challenges}
Errors occur in nested and conditional logic, such as incorrect use of \texttt{WITH} clauses \textit{"Which event had the shortest duration between P. V. Sindhu’s medal wins?"} or failing to respect conditions \textit{"List all tournaments where Carolina Marín won a medal after 2015?"}. The model also mishandles multi-field responses \textit{"List the medal type, location, and year for Hugo Calderano’s wins."}.

\vspace{-0.3em}
\paragraph{4. Aggregates, Joins, and Dependencies}
The model struggles with nested aggregates, non-standard joins, and dependency tracking. It fails to construct valid joins \textit{"Which format had the highest number of gold medals in 2020?"} or align dependencies in complex queries \textit{"Which medal did Michael Phelps win in the same tournament as his fastest recorded swim?"}. It also ignores group-level constraints, leading to overgeneralized results \textit{"Which city hosted the most gold-medal-winning tournaments for P. V. Sindhu?"}.

\vspace{-0.3em}
\paragraph{5. Inconsistencies and Robustness Issues}
Inconsistent query structures lead to variable results for similar tasks \textit{"How old was Hugo Calderano when he won his first medal?"} vs. \textit{"At what age did Michael Phelps win his most recent Olympic Gold Medal?"}. The model struggles with entity disambiguation \textit{"List all the medals won by Michael Phelps in the Olympic Games?"} and overlooks edge cases \textit{"How many medals has an athlete with no wins received?"}. Ranking logic is often mishandled, such as ignoring ordering requirements \textit{"Which city hosted the most tournaments in 2019?"}.

\section{Comparison with Related Work}

Temporal reasoning in LLMs is an evolving field intersecting with advancements in tabular reasoning, logic, and symbolic methods. Our work advances this area by introducing the \datasetName dataset for detailed evaluation of temporal reasoning in tabular contexts. Key advancements in related areas are discussed below.

\paragraph{Tabular Reasoning.}
Applying LLMs to semi-structured tables has been studied for question answering, semantic parsing, and table-to-text generation \cite{Chen2020TabFact:, gupta-etal-2020-infotabs,Zhang:2020:SET, Zhang:2020:WTE}. Approaches like TAPAS \cite{Herzig_2020}, TaBERT \cite{yin2020tabert}, and TABBIE \cite{iida2021tabbie} enhance table understanding through joint tabular-textual embeddings, while Table2vec \cite{Zhang_2019} and TabGCN \cite{pramanick-bhattacharya-2021-joint} investigate alternative representations.

Recent work also explores symbolic reasoning for structured tables with fixed schemas \cite{cheng2023bindinglanguagemodelssymbolic, ye2023largelanguagemodelsversatile, wang2024chainoftableevolvingtablesreasoning}. Building on these advances, we use SQL-based symbolic methods for temporal queries in semi-structured data. Our \datasetName dataset further enables fine-grained evaluation of temporal reasoning across diverse tabular characteristics.

\paragraph{Temporal Reasoning.} 
Temporal reasoning is central to question answering and event-centric tasks, with datasets like TIME-SENSITIVEQA \cite{chen2021a} and TORQUE \cite{ning-etal-2020-torque} addressing time-sensitive comprehension in text, and TEMPQA-WD \cite{neelam2022benchmark} and CRONQUESTIONS \cite{saxena-etal-2021-question} focusing on temporal links in knowledge graphs. Models like CRONKBQA \cite{saxena-etal-2021-question} further enhance performance by incorporating temporal reasoning during training.

Our work extends these efforts to structured tabular datasets. While datasets such as TempTabQA \cite{gupta-etal-2023-temptabqa} and TRAM \cite{wang2024tram} tackle similar challenges, \datasetName advances the field by introducing counterfactual reasoning, scalable table sizes, and diverse question difficulties, offering a broader framework for evaluating temporal reasoning.

\paragraph{Logical Reasoning and Symbolic Approaches}
Frameworks like LOGIC-LM \cite{LOGIC_LM2023} and neurosymbolic methods such as LINC \cite{LINC2023} show that integrating symbolic reasoning and external tools enhances logical inference in LLMs. Auto-formalization (e.g., NL2FOL \cite{NL2FOL2023}) further boosts reasoning accuracy by mapping natural language to structured forms.

Building on these advances, our SQL-based symbolic approach enables precise temporal reasoning over tables by translating queries into executable SQL. The \datasetName dataset offers a comprehensive benchmark at the intersection of tabular, temporal, and symbolic reasoning, featuring original and counterfactual splits, scalable table sizes, and varied question difficulties—all aligned with SQL-based structured reasoning.

\section{Conclusion}
This work investigates temporal tabular question answering with LLMs, tackling key challenges in counterfactual robustness, data sensitivity, and question complexity. We introduced \datasetName, a controlled benchmark designed for systematic evaluations. By combining symbolic intermediate representations with adaptive few-shot prompting, our approach leverages database schemas and SQL query generation to address the limitations of direct prompting.

Our experiments demonstrate that symbolic representations improve generalization, counterfactual robustness, and scalability, especially when handling larger tables. Additionally, adaptive prompting enhances reasoning for complex queries. These results highlight the necessity for the model to be data-blind, reasoning exclusively on metadata—such as the table schema—to ensure robust reasoning rather than relying on mere memorization. Future work can focus on conducting detailed error analysis and exploring fine-tuning techniques. A deeper analysis of the results will further illuminate the strengths and limitations of the approach. \datasetName lays a strong foundation for advancing structured temporal reasoning in LLMs and encourages future efforts to develop interpretable and robust temporal reasoning in AI systems.

\section*{Limitations}
We demonstrated the effectiveness of our approach through extensive experiments in English. However, extending the study to a multilingual context could reveal its applicability across diverse languages. While our work focuses on simple, entity-centric tables, real-world datasets are often more complex, such as hierarchical or multi-relational tables. Future research should explore these more intricate structures to expand the method's utility.

Our experiments assume static tables, yet many real-world scenarios involve dynamic data, such as streaming or frequently updated tables. Adapting the method to handle evolving datasets would enhance its practical relevance. Additionally, the approach does not leverage external domain knowledge, which could complement symbolic reasoning and broaden its applications.

The dataset may also exhibit inherent biases, such as domain-specific or entity-centric constraints, limiting generalizability. Future datasets should aim for greater diversity to better reflect real-world scenarios. Finally, due to computational constraints, we did not fine-tune models on the \datasetName dataset. Future work should address this limitation by exploring fine-tuning on larger datasets and evaluating the approach in more resource-intensive and dynamic settings for a comprehensive assessment.

\section*{Ethics Statement}

We are deeply committed to upholding the highest ethical standards in research and publication. To ensure transparency and reproducibility, we will publicly release our code, enhanced evaluation set, and detailed documentation, enabling the research community to validate, reproduce, and build upon our work. By sharing our resources, we aim to foster collaboration and accountability within the computational linguistics field.

Our methodology reflects a commitment to the responsible and fair use of tools and techniques, with all claims grounded in rigorously validated experimental results. To address the stochastic nature of black-box models, we maintained a fixed temperature throughout our experiments, ensuring consistent outcomes. AI tools were employed responsibly during the writing process, with careful oversight to prevent bias or inaccuracies. We provide comprehensive details about annotations, dataset splits, models, and prompting methods to ensure full reproducibility and empower researchers to evaluate our work rigorously.

Recognizing the importance of inclusivity and fairness, we acknowledge that our dataset may carry inherent biases, such as domain-specific or entity-centric limitations. While we strive for broad applicability, future iterations will prioritize greater diversity to enhance fairness and generalizability. By adhering to these principles, we aim to advance knowledge in computational linguistics while promoting ethical and responsible research practices that emphasize transparency, equity, and reproducibility.

\section*{Acknowledgments}
This work was primary sponsored by the Army Research Office under Grant Number W911NF-20-1-0080. The views and conclusions expressed herein are those of the authors and do not necessarily reflect the official policies, either expressed or implied, of the Army Research Office or the U.S. Government. The U.S. Government is authorized to reproduce and distribute reprints for governmental purposes notwithstanding any copyright notice herein. This research was also partially supported by ONR Contract N00014-19-1-2620. V.S. acknowledges support in part from NSF awards \#2217154 and \#2411319.

We are grateful to the Utah NLP group (University of Utah) and the CogComp group (University of Pennsylvania) for their valuable insights. We also thank Dibyakanti Kumar and Harsh Kumar for their significant contributions to the development of the dataset. The author is also grateful to Prof. Ana Marasović, University of Utah for their support. We appreciate the thoughtful feedback from our anonymous reviewers and the computational support provided by the Complex Data Analysis and Reasoning Lab at Arizona State University. Lastly, we thank our CoRAL lab cat, Coco, for her unwavering dedication to ensuring that professor, V.G., remained just the right amount of creatively unhinged during deadline season.

\bibliography{anthology, custom}

\section{Appendix}
\subsection{Examples:}
\subsubsection{Example 1:}

\noindent\textit{Q. Which Olympic year marked Michael Phelps' record for the most gold medals won?\\}

\noindent\textbf{Steps for SQL Reasoning}
\vspace{0.25cm}

\noindent\textbf{Step 1:} Start with the infobox table of Michael Phelps' medals

\begin{center}
\vspace{-0.0cm}
\footnotesize
\begin{tabular}{c|c|c}
\hline
\textbf{Medal} & \textbf{Year} & \textbf{Event} \\
\hline
Gold & 2008 Beijing & 100 m butterfly \\
Gold & 2008 Beijing & 200 m medley \\
Gold & 2004 Indianapolis & 200 m freestyle \\
Silver & 2002 Yokohama & 4\texttimes200 m freestyle \\
\hline
\end{tabular}

\vspace{-0.0cm}
\end{center}

\vspace{-0.0cm}
\noindent\textbf{Step 2:} Transform the data (all swimmer infoxes) into a relational schema and organize it into structured database tables for efficient querying. \\

\begin{center}
\vspace{-0.50cm}
    \noindent\textbf{Database Schema:}
\vspace{-0.25cm}
\end{center}

{ \footnotesize 
\begin{verbatim}
Athlete Table:
+-------------+----------------+
| Column      | Description    |
+-------------+----------------+
| athlete_id  | Primary Key    |
| name        | Athlete Name   |
+-------------+----------------+
Tournament Table:
+----------------+------------------------+
| Column         | Description            |
+----------------+------------------------+
| tournament_id  | Primary Key            |
| athlete_id     | Foreign Key (Athlete)  |
| name           | Tournament Name        |
+----------------+------------------------+
Format Table:
+----------------+----------------------------+
| Column         | Description                |
+----------------+----------------------------+
| format_id      | Primary Key                |
| tournament_id  | Foreign Key (Tournament)   |
| name           | Event Name                 |
+----------------+----------------------------+
Medal Table:
+-------------+----------------------------+
| Column      | Description                |
+-------------+----------------------------+
| medal_id    | Primary Key                |
| format_id   | Foreign Key (Format)       |
| type        | Medal Type                 |
| year        | Year of Achievement        |
| location    | Medal Location             |
+-------------+----------------------------+
PersonalInformation Table:
+----------------+-----------------------------+
| Column         | Description                 |
+----------------+-----------------------------+
| info_id        | Primary Key                 |
| athlete_id     | Foreign Key (Athlete)       |
| birth_year     | Birth Year                  |
| birth_month    | Birth Month                 |
| birth_day      | Birth Day                   |
+----------------+-----------------------------+

\end{verbatim}
}

\vspace{0.5cm}
\noindent\textbf{Step 3: Write the SQL Query}

\noindent The following query retrieves the year with the most gold medals:

\vspace{0.3cm}
\begin{lstlisting}[language=SQL]
WITH gold_medal_counts AS (
    SELECT m.year, COUNT(m.medal_id) AS gold_medals
    FROM Medal m
    JOIN Format f ON m.format_id = f.format_id
    JOIN Tournament t ON f.tournament_id = t.tournament_id
    JOIN Athlete a ON t.athlete_id = a.athlete_id
    WHERE a.name = 'Michael Phelps'
      AND m.type = 'MedalGold'
    GROUP BY m.year
)
SELECT year
FROM gold_medal_counts
WHERE gold_medals = (
    SELECT MAX(gold_medals)
    FROM gold_medal_counts
);
\end{lstlisting}

\vspace{0.5cm}
\noindent\textbf{Step 4: Execute the Query}

\noindent The query outputs the year with the highest number of gold medals.\\
\noindent\textbf{Final Result:} \textcolor{green!40!black!60}{\textbf{2008}}

\vspace{0.5cm}
\noindent\textbf{Direct Reasoning with Chain-of-Thought (CoT):}

\noindent To perform direct reasoning using Chain-of-Thought (CoT), LLM arrange the medal in year and count the number of gold medals per year from the table:

{\footnotesize

\begin{verbatim}
Year 2008:
- 100 m butterfly (Gold)
Total: 1 gold medals

Year 2004:
- 200 m freestyle (Gold)
- 200 m medley (Gold)
Total: 2 gold medal

Year 2002:
- 4x200 m freestyle (Silver)
Total: 0 gold medals
\end{verbatim}}

\vspace{-0.5em}
\noindent \textbf{Answer (CoT Reasoning):} \textcolor{red}{2004} has the most gold medals with a count of 2.

However, due to direct reasoning errors or omissions, it misinterpret the complex table, and hence CoT fails whereas Symbolic succeed.

\subsubsection{Example 2:}
\noindent\textit{Q. Does Emma Weyant have more Bronze Medals than Gold Medals ?\\}

\noindent\textbf{Steps for SQL Reasoning}
\vspace{0.25cm}

\noindent\textbf{Step 1:} Start with the infobox table of Emma Weyant's medals.

\begin{figure}[h]
    \centering
    \includegraphics[width=0.4\textwidth]{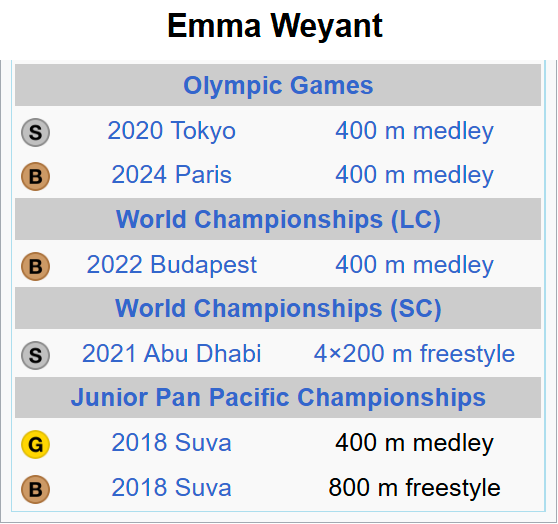}
    \caption{Emma Weyant's Medal Infobox}
    \label{fig:emma_infobox}
\end{figure}

\vspace{-0.0cm}
\noindent\textbf{Step 2:} Transform the data (all swimmer infoboxes) into a relational schema and organize it into structured database tables for efficient querying. \\

\begin{center}
\vspace{-0.50cm}
    \noindent\textbf{Database Schema:}
\vspace{-0.25cm}
\end{center}

{ \footnotesize 
\begin{verbatim}
Athlete Table:
+-------------+----------------+
| Column      | Description    |
+-------------+----------------+
| athlete_id  | Primary Key    |
| name        | Athlete Name   |
+-------------+----------------+

Tournament Table:
+----------------+------------------------+
| Column         | Description            |
+----------------+------------------------+
| tournament_id  | Primary Key            |
| athlete_id     | Foreign Key (Athlete)  |
| name           | Tournament Name        |
+----------------+------------------------+

Format Table:
+----------------+----------------------------+
| Column         | Description                |
+----------------+----------------------------+
| format_id      | Primary Key                |
| tournament_id  | Foreign Key (Tournament)   |
| name           | Event Name                 |
+----------------+----------------------------+

Medal Table:
+-------------+----------------------------+
| Column      | Description                |
+-------------+----------------------------+
| medal_id    | Primary Key                |
| format_id   | Foreign Key (Format)       |
| type        | Medal Type                 |
| year        | Year of Achievement        |
| location    | Medal Location             |
+-------------+----------------------------+

PersonalInformation Table:
+----------------+-----------------------------+
| Column         | Description                 |
+----------------+-----------------------------+
| info_id        | Primary Key                 |
| athlete_id     | Foreign Key (Athlete)       |
| birth_year     | Birth Year                  |
| birth_month    | Birth Month                 |
| birth_day      | Birth Day                   |
+----------------+-----------------------------+
\end{verbatim}
}

\vspace{0.5cm}
\noindent\textbf{Step 3: Write the SQL Query}

\noindent The following query checks whether Emma Weyant has more Bronze medals than Gold medals:

\vspace{0.3cm}
\begin{lstlisting}[language=SQL]
SELECT CASE 
    WHEN SUM(CASE WHEN m.type = 'MedalBronze' THEN 1 ELSE 0 END) > 
         SUM(CASE WHEN m.type = 'MedalGold' THEN 1 ELSE 0 END)
    THEN 'Yes'
    ELSE 'No'
END AS has_more_bronze_than_gold
FROM Medal m
JOIN Format f ON m.format_id = f.format_id
JOIN Tournament t ON f.tournament_id = t.tournament_id
JOIN Athlete a ON t.athlete_id = a.athlete_id
WHERE a.name = 'Emma Weyant';
\end{lstlisting}

\vspace{0.5cm}
\noindent\textbf{Step 4: Execute the Query}

\noindent The query outputs whether Emma Weyant has more Bronze medals than Gold medals.\\
\noindent\textbf{Final Result:} \textcolor{green!40!black!60}{\textbf{Yes}}

\vspace{0.5cm}
\noindent\textbf{Direct Reasoning with Chain-of-Thought (COT):}

\noindent Using manual reasoning, the LLM counts the medals directly from the table:

{\footnotesize
\begin{verbatim}
Gold Medals:
- 2018 Suva: 400 m medley
Total: 1 Gold Medal

Bronze Medals:
- 2018 Suva: 800 m freestyle
- 2022 Budapest: 400 m medley
- 2024 Paris: 400 m medley
Total: 3 Bronze Medals

Final Count:
Gold: 1
Bronze: 3
\end{verbatim}
}

\noindent \textbf{LLM's Answer:} \textcolor{red}{No} \& Emma Weyant has one Gold Medal and three Bronze Medals.

\vspace{0.5cm}
\noindent\textbf{Why the LLM's Answer is Incorrect and Symbolic Reasoning Succeeds:}
\begin{itemize}
    \item \textbf{Direct Reasoning Errors}: The LLM correctly identifies the count but fails in its logical comparison, leading to an incorrect conclusion.
    \item \textbf{Symbolic Reasoning Accuracy}: SQL-based reasoning explicitly performs the correct comparison and produces an unambiguous result.
    \item \textbf{Scalability and Consistency}: SQL-based methods remain reliable as data size and complexity grow, unlike manual reasoning.
\end{itemize}
\noindent\textbf{Conclusion:} Symbolic SQL reasoning eliminates errors inherent in manual reasoning methods like Chain-of-Thought, ensuring precise and reliable results.

\subsubsection{Example 3:}

\noindent\textit{Q. In which city did Yohan Blake win his first medal?}

\vspace{0.5cm}
\noindent\textbf{Step 1:} Start with the infobox table of Yohan Blake's medals.

\begin{center}
\vspace{-0.0cm}
\footnotesize
\textbf{Yohan Blake's Medal Record:}
\begin{tabular}{l|l|l}
\hline
\multicolumn{3}{c}{\textbf{Olympic Games}} \\
\hline
\textbf{Medal} & \textbf{Year} & \textbf{Format} \\
\hline
Gold & 2012 London & 4$\times$100 m relay \\
Gold & 2016 Rio de Janeiro & 4$\times$100 m relay \\
Silver & 2012 London & 100 m \\
Silver & 2012 London & 200 m \\
\hline
\multicolumn{3}{c}{\textbf{World Championships}} \\
\hline
Gold & 2011 Daegu & 100 m \\
Gold & 2011 Daegu & 4$\times$100 m relay \\
\hline
\multicolumn{3}{c}{\textbf{Commonwealth Games}} \\
\hline
Bronze & 2018 Gold Coast & 100 m \\
Bronze & 2018 Gold Coast & 4$\times$100 m relay \\
\hline
\multicolumn{3}{c}{\textbf{World Relays}} \\
\hline
Gold & 2014 Bahamas & 4$\times$100 m \\
Gold & 2014 Bahamas & 4$\times$200 m \\
Bronze & 2017 Bahamas & 4$\times$200 m \\
\hline
\multicolumn{3}{c}{\textbf{World Junior Championships}} \\
\hline
Gold & 2006 Beijing & 4$\times$100 m relay \\
Silver & 2008 Bydgoszcz & 4$\times$100 m relay \\
Bronze & 2006 Beijing & 100 m \\
\hline
\multicolumn{3}{c}{\textbf{Pan American Junior Championships}} \\
\hline
Silver & 2007 São Paulo & 100 m \\
Bronze & 2007 São Paulo & 4$\times$400 m relay \\
\hline
\multicolumn{3}{c}{\textbf{CAC Junior Championships (U20)}} \\
\hline
Gold & 2006 Port of Spain & 100 m \\
Gold & 2006 Port of Spain & 200 m \\
Gold & 2006 Port of Spain & 4$\times$100 m relay \\
\hline
\multicolumn{3}{c}{\textbf{CARIFTA Games}} \\
\hline
Gold & 2006 Les Abymes & 200 m \\
Gold & 2006 Les Abymes & 4$\times$100 m relay \\
Gold & 2007 Providenciales & 100 m \\
Gold & 2007 Providenciales & 4$\times$100 m relay \\
Gold & 2008 Basseterre & 100 m \\
\hline
\multicolumn{3}{c}{\textbf{CARIFTA Games}} \\
\hline
Gold & 2005 Bacolet & 100 m \\
Gold & 2005 Bacolet & 200 m \\
\hline
\multicolumn{3}{c}{\textbf{Continental Cup}} \\
\hline
Gold & 2018 Ostrava & 4$\times$100 m \\
\hline
\end{tabular}
\vspace{-0.0cm}
\end{center}

\vspace{-0.0cm}
\noindent\textbf{Step 2:} Transform the data into a relational schema and organize it into structured database tables for efficient querying (similar to Step 2 in previous examples). 

\vspace{0.5cm}
\noindent\textbf{Step 3: Write the SQL Query}

\noindent The following query retrieves the location where Yohan Blake won his first medal:

\begin{lstlisting}[language=SQL]
SELECT DISTINCT m.location
FROM Medal m
JOIN Format f ON m.format_id = f.format_id
JOIN Tournament t ON f.tournament_id = t.tournament_id
JOIN Athlete a ON t.athlete_id = a.athlete_id
WHERE a.name = 'Yohan Blake'
  AND m.year = (
    SELECT MIN(m2.year)
    FROM Medal m2
    JOIN Format f2 ON m2.format_id = f2.format_id
    JOIN Tournament t2 ON f2.tournament_id = t2.tournament_id
    JOIN Athlete a2 ON t2.athlete_id = a2.athlete_id
    WHERE a2.name = 'Yohan Blake'
  );
\end{lstlisting}

\vspace{0.5cm}
\noindent\textbf{Step 4: Execute the Query}

\noindent The query outputs the location where Yohan Blake won his first medal.\\
\noindent\textbf{Final Result:} \textcolor{green!40!black!60}{\textbf{Bacolet}}

\vspace{0.5cm}
\noindent\textbf{Direct Reasoning with Chain-of-Thought (COT):}

\noindent Using manual reasoning, the LLM incorrectly identifies the location as Beijing:

{\footnotesize
\begin{verbatim}
Year 2006:
- Gold: 4x100 m relay (World Junior Championships, Beijing)
- Bronze: 100 m (World Junior Championships, Beijing)
Conclusion: First medal location is Beijing.
\end{verbatim}
}

\noindent \textbf{LLM's Answer:} \textcolor{red}{Beijing}. In 2006, Yohan Blake won his first medal at the World Junior Championships in Beijing, where he secured a Gold in the 4x100 m relay and a Bronze in the 100 m.

\vspace{0.5cm}
\noindent\textbf{Why the LLM's Answer is Incorrect and Symbolic Reasoning Succeeds:}
\begin{itemize}
    \item \textbf{Direct Reasoning Errors}: The LLM overlooks earlier results from 2005 in the CARIFTA Games held in Bacolet, where Yohan Blake won two Gold medals.
    \item \textbf{Symbolic Reasoning Accuracy}: SQL explicitly finds the minimum year and correctly identifies the location associated with the first medal.
    \item \textbf{Consistency and Scalability}: Symbolic SQL reasoning reliably handles large, complex medal records without omission or error.
\end{itemize}

\noindent\textbf{Conclusion:} Symbolic SQL reasoning eliminates the errors inherent in Chain-of-Thought reasoning, ensuring accurate and reliable results.

\subsubsection{Example 4:}
\noindent\textit{Q. How many medals did Mayu Matsumoto win in her twenties?}

\vspace{0.5cm}
\noindent\textbf{Step 1:} Start with the infobox table of Mayu Matsumoto's medals.

\begin{figure}[h]
    \centering
    \includegraphics[width=0.3\textwidth]{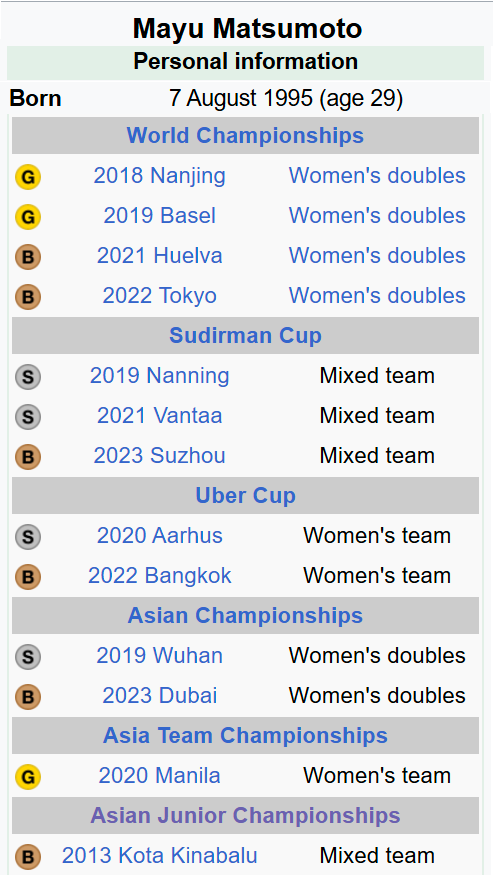}
    \caption{Mayu Matsumoto's Medal Infobox}
    \label{fig:mayu_infobox}
\end{figure}

\vspace{-0.0cm}
\noindent\textbf{Step 2:} Transform the data into a relational schema and organize it into structured database tables for efficient querying. (similar to Step 2 in previous examples).

\vspace{0.5cm}
\noindent\textbf{Step 3: Write the SQL Query}

\noindent The following query calculates how many medals Mayu Matsumoto won between the ages of 20 and 29:

\begin{lstlisting}[language=SQL]
SELECT COUNT(m.medal_id) AS total_medals
FROM Medal m
JOIN Format f ON m.format_id = f.format_id
JOIN Tournament t ON f.tournament_id = t.tournament_id
JOIN Athlete a ON t.athlete_id = a.athlete_id
JOIN PersonalInformation pi ON a.athlete_id = pi.athlete_id
WHERE a.name = 'Mayu Matsumoto'
  AND (m.year - pi.birth_year) BETWEEN 20 AND 29;
\end{lstlisting}

\vspace{0.5cm}
\noindent\textbf{Step 4: Execute the Query}

\noindent The query calculates medals for Mayu Matsumoto won during her twenties (from 2015 to 2024, since she was born in 1995). \\

\noindent\textbf{Final Result:} \textcolor{green!40!black!60}{\textbf{12}}

\vspace{0.5cm}
\noindent\textbf{Direct Reasoning with Chain-of-Thought (COT):}

\noindent Using manual reasoning, the LLM incorrectly calculates the total as follows:

\noindent Mayu Matsumoto, born in \textbf{1995}, was in her twenties from \textbf{2015 to 2024}. During this period:

\noindent \textbf{World Championships:} 2 Gold (2018, 2019), 1 Silver (2021), 2 Bronze (2021, 2022).

\noindent \textbf{Sudirman Cup:} 2 Silver (2019, 2021), 1 Bronze (2023).

\noindent \textbf{Uber Cup:} 1 Silver (2020), 1 Bronze (2022).

\noindent \textbf{Asian Championships:} 1 Silver (2019), 1 Bronze (2023).

\noindent Total (COT Reasoning): 10 medals.

\vspace{0.5cm}
\noindent \textbf{Trends in the LLM's Reasoning:}
\begin{itemize}
    \item \textbf{Incorrect Medal Count for World Championships:} The LLM claims 2 Gold, 2 Silver, and 1 Bronze, but the infobox shows 2 Gold (2018, 2019), 1 Bronze (2022), and only 1 Silver (2021).
    \item \textbf{Overcounted/Undercounted Totals:} The total medals, when carefully counted, sum to 12, not 10:
    \begin{itemize}
        \item \textbf{World Championships:} 2 Gold, 1 Silver, 1 Bronze (Total = 4).
        \item \textbf{Sudirman Cup:} 2 Silver, 1 Bronze (Total = 3).
        \item \textbf{Uber Cup:} 1 Silver, 1 Bronze (Total = 2).
        \item \textbf{Asian Championships:} 1 Silver, 1 Bronze (Total = 2).
        \item \textbf{Incorrectly Excluded 2020 Medal:} Asian Team Championships (2020, age 25) is excluded incorrectly.
        \item \textbf{Correctly Excluded 2013 Medal:} Asian Junior Championships (2013, age 18) is excluded correctly.
        
    \end{itemize}
    \item \textbf{Temporal Misinterpretation:} The LLM fails to count some of the medals in the 20-29 age range and fails to sum them accurately.
\end{itemize}

\vspace{0.5cm}
\noindent\textbf{Symbolic Reasoning Accuracy:}
\begin{itemize}
    \item SQL precisely filters years between 2015 and 2024, ensuring only valid medals are counted.
    \item Symbolic reasoning eliminates human counting errors and temporal miscalculations.
    \item The result is accurate: \textbf{12 medals}.
\end{itemize}

\vspace{0.5cm}
\noindent\textbf{Conclusion:} The LLM's Chain-of-Thought reasoning undercounts Mayu Matsumoto's medals, providing an incorrect total of 10 due to miscounting and temporal errors. Symbolic SQL reasoning accurately identifies the correct total as \textcolor{green!40!black!60}{\textbf{12}} medals won during her twenties.

\subsubsection{Example 5:}
\noindent\textit{Q. How many times did Sandra Sánchez win a medal in the World Championships before 2021?}

\vspace{0.5cm}
\noindent\textbf{Step 1:} Start with the infobox table of Sandra Sánchez's medals.

\begin{figure}[h]
    \centering
    \includegraphics[width=0.3\textwidth]{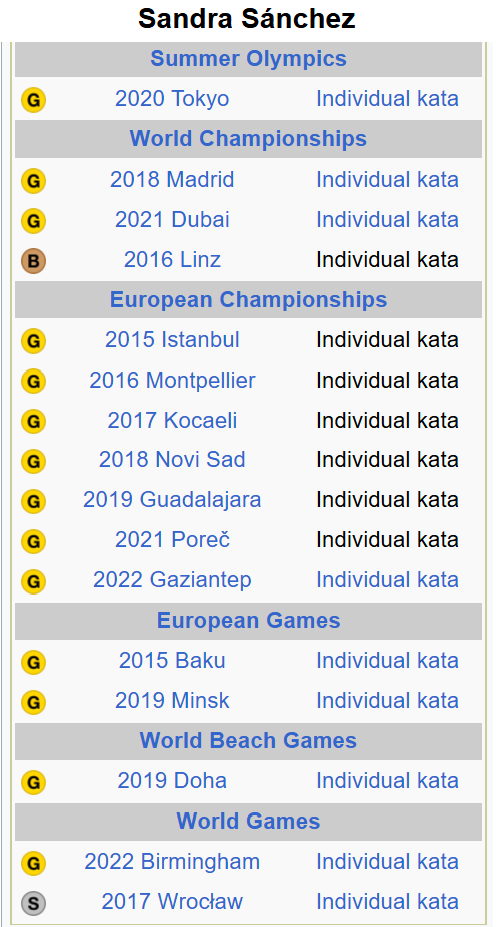}
    \caption{Sandra Sánchez's Medal Infobox}
    \label{fig:sandra_infobox}
\end{figure}

\vspace{-0.0cm}
\noindent\textbf{Step 2:} Transform the data into a relational schema and organize it into structured database tables for efficient querying (similar to Step 2 in previous examples).

\vspace{0.5cm}
\noindent\textbf{Step 3: Write the SQL Query}

\noindent The following query calculates how many medals Sandra Sánchez won in the World Championships before the year 2021:

\begin{lstlisting}[language=SQL]
SELECT COUNT(m.medal_id) AS total_medals
FROM Medal m
JOIN Format f ON m.format_id = f.format_id
JOIN Tournament t ON f.tournament_id = t.tournament_id
JOIN Athlete a ON t.athlete_id = a.athlete_id
WHERE a.name = 'Sandra Sánchez'
  AND t.name = 'World Championships'
  AND m.year < 2021;
\end{lstlisting}

\vspace{0.5cm}
\noindent\textbf{Step 4: Execute the Query}

\noindent The query outputs the total number of medals Sandra Sánchez won in the World Championships before 2021.\\

\noindent\textbf{Final Result:} \textcolor{green!40!black!60}{\textbf{2}}

\vspace{0.5cm}
\noindent\textbf{Direct Reasoning with Chain-of-Thought (COT):}

\noindent The LLM incorrectly provides the following reasoning:

{\footnotesize
\noindent Sandra Sánchez won a Bronze medal in the World Championships in 2016, which is before 2021. Therefore, the answer is 1.
}

\vspace{0.5cm}
\noindent\textbf{Errors in the LLM's Reasoning:}
\begin{itemize}
    \item \textbf{Missed Medal in 2018:} While the LLM identifies the 2016 Bronze medal, it fails to recognize the 2018 Gold medal in Madrid, which also occurred before 2021.
    \item \textbf{Incomplete Temporal Analysis:} The LLM does not account for all relevant years when performing temporal reasoning, leading to an undercount of medals.
\end{itemize}

\vspace{0.5cm}
\noindent\textbf{Symbolic Reasoning Accuracy:}
\begin{itemize}
    \item SQL explicitly filters medals in the World Championships where the year is less than 2021.
    \item The query correctly identifies both the 2016 Bronze medal and the 2018 Gold medal, producing the accurate total of \textbf{2 medals}.
    \item Symbolic reasoning eliminates human oversight by systematically querying all relevant data within the temporal range.
\end{itemize}

\vspace{0.5cm}
\noindent\textbf{Conclusion:} The LLM's Chain-of-Thought reasoning incorrectly counts only \textbf{1 medal} due to missed temporal filtering. Symbolic SQL reasoning, by explicitly querying for medals before 2021, produces the correct result: \textcolor{green!40!black!60}{\textbf{2 medals}}.

\subsection{Result Analysis for all models:}

\subsubsection{Analysis for GPT-4o:}
\noindent From Table~\ref{tab:gpt4o_results}, comparing \textbf{SQL Adaptive} with \textbf{Direct Adaptive} across key aspects, we observe:

\begin{itemize}
\item \textbf{Counterfactual Gap:}
For \textbf{Table - Adaptive} (EMS), the gap between Original (\textbf{58.13}) and CounterFact (\textbf{40.92}) is \textbf{17.21}.
For \textbf{SQL Schema - Adaptive} (EMS), the gap reduces to \textbf{2.96}, indicating improved robustness to counterfactual data.

\item \textbf{Scalability to Table Size:}  
For \textbf{Table - Adaptive} (EMS), the gap between Large (\textbf{48.88}) and Small (\textbf{73.92}) tables is \textbf{25.04}.  
For \textbf{SQL Schema - Adaptive} (EMS), the gap decreases significantly to \textbf{1.06}, demonstrating better scalability to large table sizes.

\item \textbf{Question Complexity:}  
For \textbf{Table - Adaptive} (EMS), the performance on Easy, Medium, and Hard questions is \textbf{74.38}, \textbf{63.91}, and \textbf{54.17}, respectively.  
For \textbf{SQL Schema - Adaptive} (EMS), the performance improves to \textbf{80.06} (Easy), \textbf{73.37} (Medium), and \textbf{66.74} (Hard), showing better handling of increasing question complexity.

\item \textbf{Adaptive Few-Shot Effectiveness:}  
For Original data, \textbf{SQL Schema - Adaptive} (EMS: \textbf{71.63}) outperforms \textbf{Table - Adaptive} (EMS: \textbf{58.13}), highlighting the benefit of adaptive prompting in achieving higher accuracy.
\end{itemize}

\noindent These results demonstrate that \textbf{SQL Adaptive} consistently outperforms \textbf{Direct Adaptive} by reducing the counterfactual gap, improving scalability to large tables, and enhancing performance across question complexities.

\subsubsection{Analysis for GPT-4o Mini:}
\noindent From Table~\ref{tab:gpt_4o_mini_results}, comparing \textbf{SQL Adaptive} with \textbf{Table Adaptive} across key aspects, we observe:

\begin{itemize}
    \item \textbf{Counterfactual Gap:}  
    For \textbf{Table - Adaptive} (EMS), the gap between Original (\textbf{48.79}) and CounterFact (\textbf{35.48}) is \textbf{13.31}.  
    For \textbf{SQL Schema - Adaptive} (EMS), the gap reduces significantly to \textbf{3.45} (Original: \textbf{68.69}, CounterFact: \textbf{65.24}), indicating improved robustness to counterfactual data.

    \item \textbf{Scalability to Table Size:}  
    For \textbf{Table - Adaptive} (EMS), the gap between Large (\textbf{39.96}) and Small (\textbf{64.80}) tables is \textbf{24.84}.  
    For \textbf{SQL Schema - Adaptive} (EMS), the gap reduces to \textbf{3.34} (Large: \textbf{65.43}, Small: \textbf{68.77}), demonstrating better scalability to large table sizes.

    \item \textbf{Question Complexity:}  
    For \textbf{Table - Adaptive} (EMS), the scores for Easy, Medium, and Hard questions are \textbf{63.16}, \textbf{52.07}, and \textbf{44.49}, respectively.  
    For \textbf{SQL Schema - Adaptive} (EMS), the performance improves to \textbf{76.87} (Easy), \textbf{70.41} (Medium), and \textbf{63.13} (Hard), showcasing better handling of increasing question complexity.

    \item \textbf{Adaptive Few-Shot Effectiveness:}  
    For Original data, \textbf{SQL Schema - Adaptive} (EMS: \textbf{68.69}) significantly outperforms \textbf{Table - Adaptive} (EMS: \textbf{48.79}), highlighting the superior accuracy achieved with symbolic reasoning and adaptive prompting.
\end{itemize}

\noindent These results clearly show that \textbf{SQL Adaptive} consistently outperforms \textbf{Table Adaptive}, with smaller counterfactual and table size gaps, and better performance across question complexity levels.

\subsubsection{Analysis for Gemini 1.5 Flash:}
\noindent From Table~\ref{tab:gemini_1_5_flash_results}, comparing \textbf{SQL Adaptive} with \textbf{Table Adaptive}, we observe:

\begin{itemize}
    \item \textbf{Counterfactual Gap:}  
    For \textbf{Table - Adaptive} (EMS), the gap between Original (\textbf{52.90}) and CounterFact (\textbf{42.91}) is \textbf{9.99}.  
    In comparison, for \textbf{SQL Schema - Adaptive} (EMS), the gap is reduced to \textbf{2.78} (Original: \textbf{65.49}, CounterFact: \textbf{62.71}), indicating significantly improved robustness to counterfactual data.

    \item \textbf{Scalability to Table Size:}  
    For \textbf{Table - Adaptive} (EMS), the gap between Large (\textbf{41.02}) and Small (\textbf{66.25}) tables is \textbf{25.23}.  
    For \textbf{SQL Schema - Adaptive} (EMS), the gap reduces to \textbf{4.23} (Large: \textbf{69.30}, Small: \textbf{73.53}), showcasing SQL's superior handling of larger tables.

    \item \textbf{Question Complexity:}  
    For \textbf{Table - Adaptive} (EMS), the scores for Easy, Medium, and Hard questions are \textbf{65.76}, \textbf{55.95}, and \textbf{45.92}, respectively.  
    For \textbf{SQL Schema - Adaptive} (EMS), the scores improve to \textbf{76.26} (Easy), \textbf{73.72} (Medium), and \textbf{63.12} (Hard), highlighting better performance as question complexity increases.

    \item \textbf{Adaptive Few-Shot Effectiveness:}  
    For Original data, \textbf{SQL Schema - Adaptive} (EMS: \textbf{65.49}) outperforms \textbf{Table - Adaptive} (EMS: \textbf{52.90}), demonstrating the effectiveness of adaptive few-shot prompting with symbolic reasoning.
\end{itemize}

\noindent These observations show that \textbf{SQL Adaptive} significantly reduces the counterfactual gap, scales better with large tables, and consistently achieves higher accuracy across all question complexities compared to \textbf{Table Adaptive}.

\subsubsection{Analysis for Gemini 1.5 Pro:}
\noindent From Table~\ref{tab:gemini_1_5_pro_results}, comparing \textbf{SQL Adaptive} with \textbf{Table Adaptive}, we observe:

\begin{itemize}
    \item \textbf{Counterfactual Gap:}  
    For \textbf{Table - Adaptive} (EMS), the gap between Original (\textbf{53.48}) and CounterFact (\textbf{44.19}) is \textbf{9.29}.  
    For \textbf{SQL Schema - Adaptive} (EMS), the gap is reduced to \textbf{0.16} (Original: \textbf{65.29}, CounterFact: \textbf{65.13}), showcasing excellent robustness to counterfactual data.

    \item \textbf{Scalability to Table Size:}  
    For \textbf{Table - Adaptive} (EMS), the gap between Large (\textbf{41.86}) and Small (\textbf{67.27}) tables is \textbf{25.41}.  
    For \textbf{SQL Schema - Adaptive} (EMS), the gap reduces significantly to \textbf{2.88} (Large: \textbf{72.43}, Small: \textbf{75.31}), demonstrating better scalability with large tables.

    \item \textbf{Question Complexity:}  
    For \textbf{Table - Adaptive} (EMS), the scores for Easy, Medium, and Hard questions are \textbf{66.26}, \textbf{56.47}, and \textbf{46.74}, respectively.  
    For \textbf{SQL Schema - Adaptive} (EMS), the scores improve to \textbf{75.86} (Easy), \textbf{71.47} (Medium), and \textbf{59.24} (Hard), highlighting superior handling of increasing question complexity.

    \item \textbf{Adaptive Few-Shot Effectiveness:}  
    For Original data, \textbf{SQL Schema - Adaptive} (EMS: \textbf{65.29}) significantly outperforms \textbf{Table - Adaptive} (EMS: \textbf{53.48}), demonstrating the clear benefits of symbolic reasoning combined with adaptive few-shot prompting.
\end{itemize}

\noindent These results clearly highlight that \textbf{SQL Adaptive} consistently reduces counterfactual gaps, scales better with table size, and improves performance across question complexities compared to \textbf{Table Adaptive}.

\subsubsection{Analysis for Llama 3.1 70B:}

\noindent From Table~\ref{tab:llama_3_1_70b_results}, comparing \textbf{SQL Adaptive} with \textbf{Table Adaptive}, we observe:

\begin{itemize}
    \item \textbf{Counterfactual Gap:}  
    For \textbf{Table - Adaptive} (EMS), the gap between Original (\textbf{53.63}) and CounterFact (\textbf{39.20}) is \textbf{14.43}.  
    For \textbf{SQL Schema - Adaptive} (EMS), the gap reduces to \textbf{0.55} (Original: \textbf{64.36}, CounterFact: \textbf{63.81}). SQL Adaptive is clearly more robust to counterfactual data.

    \item \textbf{Scalability to Table Size:}  
    For \textbf{Table - Adaptive} (EMS), the gap between Large (\textbf{46.65}) and Small (\textbf{68.07}) tables is \textbf{21.42}.  
    For \textbf{SQL Schema - Adaptive} (EMS), the gap remains smaller at \textbf{1.98} (Large: \textbf{66.91}, Small: \textbf{68.89}), showing better performance scalability with table size.

    \item \textbf{Question Complexity:}  
    For \textbf{Table - Adaptive} (EMS), the scores for Easy, Medium, and Hard questions are \textbf{69.40}, \textbf{59.76}, and \textbf{52.85}, respectively.  
    For \textbf{SQL Schema - Adaptive} (EMS), the scores are \textbf{77.19} (Easy), \textbf{67.65} (Medium), and \textbf{60.92} (Hard), showing a consistent improvement across complexities.

    \item \textbf{Adaptive Few-Shot Effectiveness:}  
    For Original data, \textbf{SQL Schema - Adaptive} (EMS: \textbf{64.36}) performs significantly better than \textbf{Table - Adaptive} (EMS: \textbf{53.63}), confirming the benefit of symbolic reasoning with adaptive few-shot prompting.
\end{itemize}

\noindent Overall, \textbf{SQL Adaptive} demonstrates clear improvements over \textbf{Table Adaptive} in counterfactual robustness, scalability to table size, and performance across question complexity levels. The observed gaps in Table Adaptive remain substantial, especially for counterfactual and large table scenarios.

\subsubsection{Analysis for Mixtral 8x7B:}
\noindent From Table~\ref{tab:mixtral_8x7b_results}, comparing \textbf{SQL} and \textbf{Table} under both \emph{Adaptive} and \emph{Static} few-shot settings, we observe:

\begin{itemize}
    \item \textbf{Counterfactual Gap:}  
    For \textbf{Table – Adaptive} (EMS), the gap between Original (\textbf{37.54}) and CounterFact (\textbf{30.62}) is \textbf{6.92}.  
    For \textbf{SQL Schema – Adaptive} (EMS), the gap is \textbf{4.20} (Original: \textbf{25.09}, CounterFact: \textbf{20.89}).  
    For \textbf{Table Text – Static} (EMS), the gap is \textbf{8.59} (Original: \textbf{48.79}, CounterFact: \textbf{40.20}).  
    For \textbf{SQL Schema – Static} (EMS), the gap is \textbf{5.82} (Original: \textbf{46.02}, CounterFact: \textbf{40.20}).  
    SQL settings therefore show smaller counterfactual gaps than their Table counterparts in both few-shot modes.

    \item \textbf{Scalability to Table Size:}  
    For \textbf{Table – Adaptive} (EMS), the gap between Large (\textbf{34.94}) and Small (\textbf{47.72}) tables is \textbf{12.78}.  
    For \textbf{SQL Schema – Adaptive} (EMS), the gap is \textbf{9.03} (Large: \textbf{24.54}, Small: \textbf{33.57}).  
    For \textbf{Table Text – Static} (EMS), the gap is \textbf{24.14} (Large: \textbf{40.89}, Small: \textbf{65.03}).  
    For \textbf{SQL Schema – Static} (EMS), the gap is \textbf{5.02} (Large: \textbf{47.96}, Small: \textbf{52.98}).  
    SQL configurations exhibit smaller size-related drops than Table configurations, particularly in the Static setting.

    \item \textbf{Question Complexity:}  
    For \textbf{Table – Adaptive} (EMS), the scores are \textbf{50.96} (Easy), \textbf{38.46} (Medium), and \textbf{35.74} (Hard).  
    For \textbf{SQL Schema – Adaptive} (EMS), the scores are \textbf{26.78} (Easy), \textbf{46.55} (Medium), and \textbf{21.56} (Hard).  
    For \textbf{Table Text – Static} (EMS), the scores are \textbf{62.30} (Easy), \textbf{53.25} (Medium), and \textbf{45.62} (Hard).  
    For \textbf{SQL Schema – Static} (EMS), the scores are \textbf{67.35} (Easy), \textbf{45.36} (Medium), and \textbf{40.61} (Hard).  
    SQL Adaptive surpasses Table Adaptive on \textit{Medium} questions only; SQL Static exceeds Table Text Static on \textit{Easy} questions but trails on Medium and Hard ones.

    \item \textbf{Few-Shot Results on Original Subset:}  
    \textbf{Table – Adaptive} achieves \textbf{37.54} EMS, while \textbf{SQL Schema – Adaptive} reaches \textbf{25.09}.  
    \textbf{Table Text – Static} records \textbf{48.79} EMS, and \textbf{SQL Schema – Static} attains \textbf{46.02}.  
    Table maintains higher absolute EMS in both few-shot modes, though the SQL--Table gap is narrower in the Static setting.
\end{itemize}

\subsubsection{Analysis for SQL Coder 70B:}
\noindent Since this is a \textbf{code-based model}, we only evaluate baselines related to \textbf{code generation} and exclude text generation baselines. From Table~\ref{tab:sql_coder_results}, we observe the following for \textbf{SQL Schema}:

\begin{itemize}
\item \textbf{Counterfactual Gap:}
For \textbf{SQL Static} (EMS), the gap between Original (\textbf{52.08}) and CounterFact (\textbf{47.93}) is \textbf{4.15}.
For \textbf{SQL Adaptive} (EMS), the gap reduces to \textbf{2.38} (Original: \textbf{55.88}, CounterFact: \textbf{53.50}), demonstrating improved robustness with adaptive few-shot prompting.

\item \textbf{Scalability to Table Size:}  
For \textbf{SQL Static} (EMS), the gap between Large (\textbf{62.28}) and Small (\textbf{59.53}) tables is \textbf{2.75}.  
For \textbf{SQL Adaptive} (EMS), the gap is slightly larger at \textbf{4.22} (Large: \textbf{63.17}, Small: \textbf{58.95}), showing minor regression in scalability.

\item \textbf{Question Complexity:}  
For \textbf{SQL Static} (EMS), the scores for Easy, Medium, and Hard questions are \textbf{77.39}, \textbf{51.19}, and \textbf{28.91}, respectively.  
For \textbf{SQL Adaptive} (EMS), the scores improve for Medium (\textbf{58.38}) and Hard (\textbf{51.74}) questions but decrease for Easy (\textbf{63.93}), indicating uneven performance gains.

\item \textbf{Overall Accuracy:}  
For Original data, \textbf{SQL Adaptive} (EMS: \textbf{55.88}) outperforms \textbf{SQL Static} (EMS: \textbf{52.08}), highlighting the effectiveness of adaptive few-shot prompting for code-specific tasks.
\end{itemize}

\noindent Overall, SQL Adaptive demonstrates improved robustness and accuracy compared to SQL Static, particularly on counterfactual and medium-complexity queries.

\subsubsection{Analysis for Code Llama 70B:}
\noindent Since this is a \textbf{code-based model}, we only evaluate baselines related to \textbf{code generation} and exclude text generation baselines. From Table~\ref{tab:code_llama_results}, we observe the following for \textbf{SQL Schema}:

\begin{itemize}
    \item \textbf{Counterfactual Gap:}  
    For \textbf{SQL Static} (EMS), the gap between Original (\textbf{15.84}) and CounterFact (\textbf{32.62}) is substantial at \textbf{16.78}, indicating performance degradation.  
    For \textbf{SQL Adaptive} (EMS), the gap reduces to \textbf{16.53} (Original: \textbf{23.53}, CounterFact: \textbf{40.06}). While there is slight improvement, the gap remains significant.

    \item \textbf{Scalability to Table Size:}  
    For \textbf{SQL Static} (EMS), the gap between Large (\textbf{29.82}) and Small (\textbf{41.64}) tables is \textbf{11.82}.  
    For \textbf{SQL Adaptive} (EMS), the gap decreases to \textbf{10.53} (Large: \textbf{37.61}, Small: \textbf{48.14}), indicating modest improvements in handling table size.

    \item \textbf{Question Complexity:}  
    For \textbf{SQL Static} (EMS), the scores for Easy, Medium, and Hard questions are \textbf{53.42}, \textbf{41.62}, and \textbf{38.94}, respectively.  
    For \textbf{SQL Adaptive} (EMS), the scores improve across all complexities to \textbf{65.16} (Easy), \textbf{50.89} (Medium), and \textbf{40.61} (Hard), showing clear improvements, particularly for Easy and Medium questions.

    \item \textbf{Overall Accuracy:}  
    For Original data, \textbf{SQL Adaptive} (EMS: \textbf{23.53}) outperforms \textbf{SQL Static} (EMS: \textbf{15.84}), demonstrating the benefits of adaptive few-shot prompting for overall accuracy.
\end{itemize}

\noindent Overall, \textbf{SQL Adaptive} shows moderate improvements over \textbf{SQL Static}, particularly in handling table size and question complexities, though counterfactual robustness remains a challenge.

\renewcommand{\arraystretch}{1.0}
\begin{table*}[h]
\centering
\tiny
\begin{tabular}{ll|c|c|ccccccccc}
\toprule
\multirow{1}{*}{\textbf{Output}} & \multirow{1}{*}{\textbf{Context}} & \multirow{1}{*}{\textbf{Method}} & \textbf{Metric} & \textbf{Original} & \textbf{CounterFact} & \shortstack{\textbf{Gap (Original} \\ \textbf{- CounterFact)}} & \textbf{Large} & \textbf{Small} & \shortstack{\textbf{Gap (Small} \\ \textbf{- Large)}} & \textbf{Easy} & \textbf{Medium} & \textbf{Hard} \\
\midrule

\multirow{27}{*}{Text} 
& \multirow{2}{*}{None} & \multirow{2}{*}{-} & REMS & 25.91 & 13.45 & 12.46 & 23.84 & 25.79 & 1.95 & 28.67 & 25.55 & 23.83 \\
& & & EMS & 25.09 & 12.16 & 12.93 & 23.05 & 24.68 & 1.63 & 27.26 & 24.06 & 22.35 \\
\cmidrule{2-13}

& \multirow{22}{*}{Table} & \multirow{2}{*}{zero shots} & REMS & 55.94 & 42.26 & 13.68 & 45.24 & 67.81 & 22.57 & 72.34 & 59.43 & 52.90 \\
& & & EMS & 53.46 & 40.06 & 13.40 & 43.31 & 64.56 & 21.25 & 69.86 & 57.20 & 49.02 \\
\cmidrule{3-13}

& & \multirow{2}{*}{Static} & REMS & 58.69 & 44.69 & 14.00 & 48.96 & 73.57 & 24.61 & 73.45 & 65.39 & 56.97 \\
& & & EMS & 56.57 & 42.35 & 14.22 & 46.84 & 71.11 & 24.27 & 71.18 & 63.12 & 53.35 \\
\cmidrule{3-13}

& & \multirow{2}{*}{Adaptive} & REMS & 60.71 & 43.10 & 17.61 & 51.41 & 76.97 & 25.56 & 76.89 & 66.00 & 58.00 \\
& & & EMS & 58.13 & 40.92 & 17.21 & 48.88 & 73.92 & 25.04 & 74.38 & 63.91 & 54.17 \\
\cmidrule{3-13}

& & Clear & REMS & 66.84 & 50.17 & 16.67 & 55.54 & 77.86 & 22.32 & 78.97 & 72.70 & 64.72 \\
& & & EMS & 65.57 & 48.21 & 17.36 & 53.53 & 76.49 & 22.96 & 76.40 & 71.99 & 62.62 \\
\cmidrule{3-13}

& & Chain Of Thought & REMS & 70.96 & 50.43 & 20.53 & 58.36 & 76.02 & 17.66 & 81.02 & 75.52 & 66.51 \\
& & & EMS & 69.90 & 48.50 & 21.40 & 56.13 & 74.15 & 18.02 & 78.43 & 75.35 & 64.06 \\
\cmidrule{3-13}

& & Plan And Solve & REMS & 70.36 & 48.62 & 21.74 & 57.14 & 76.91 & 19.77 & 80.30 & 72.30 & 62.64 \\
& & & EMS & 69.55 & 46.50 & 23.05 & 55.20 & 75.44 & 20.24 & 77.96 & 70.81 & 60.25 \\
\cmidrule{3-13}

& & Program of Thought & REMS & 59.05 & 50.28 & 8.77 & 51.66 & 64.35 & 12.69 & 72.65 & 60.85 & 52.89 \\
& & & EMS & 56.40 & 47.07 & 9.33 & 50.19 & 62.22 & 12.03 & 69.94 & 57.20 & 48.81 \\
\cmidrule{3-13}

& & Faithful CoT & REMS & 60.02 & 50.90 & 9.12 & 51.76 & 64.85 & 13.09 & 72.02 & 59.78 & 51.25 \\
& & & EMS & 57.44 & 47.78 & 9.66 & 50.19 & 62.69 & 12.50 & 69.24 & 56.02 & 47.17 \\

\midrule

\multirow{7}{*}{SQL}
& \multirow{7}{*}{Schema} & \multirow{2}{*}{zero shots} & REMS & 51.44 & 48.94 & 2.50 & 55.81 & 59.92 & 4.11 & 63.53 & 65.20 & 45.16 \\
& & & EMS & 49.31 & 46.64 & 2.67 & 53.90 & 57.43 & 3.53 & 61.22 & 63.12 & 42.02 \\
\cmidrule{3-13}

& & \multirow{2}{*}{Static} & REMS & 66.76 & 62.41 & 4.35 & 71.82 & 75.37 & 3.55 & 81.27 & 75.88 & 64.38 \\
& & & EMS & 65.22 & 60.94 & 4.28 & 70.63 & 73.57 & 2.94 & 78.89 & 75.15 & 62.31 \\
\cmidrule{3-13}

& & \multirow{2}{*}{Adaptive} & REMS & 71.87 & 69.04 & 2.83 & 73.05 & 74.55 & 1.50 & 80.17 & 74.35 & 67.38 \\
& & & EMS & 71.63 & 68.67 & 2.96 & 72.86 & 73.92 & 1.06 & 80.06 & 73.37 & 66.74 \\

\bottomrule
\end{tabular}
\vspace{-1.0em}
\caption{\small Test Set Results for GPT-4o.}
\label{tab:gpt4o_results}
\vspace{-1.0em}
\end{table*}

\renewcommand{\arraystretch}{1.0}
\begin{table*}[h]
\centering
\footnotesize
\begin{tabular}{ll|c|c|ccccccc}
\toprule
\multirow{2}{*}{\textbf{Output}} & \multirow{2}{*}{\textbf{Context}} & \multirow{2}{*}{\textbf{Few Shots}} & \textbf{Metric} & \multicolumn{7}{c}{\textbf{Results Across Categories}} \\
\cmidrule{5-11}
& & & & \shortstack{\textbf{Original}} & \shortstack{\textbf{CounterFact}} & \shortstack{\textbf{Large}} & \shortstack{\textbf{Small}} & \shortstack{\textbf{Easy}} & \shortstack{\textbf{Medium}} & \shortstack{\textbf{Hard}} \\
\midrule

\multirow{10}{*}{Text} & \multirow{2}{*}{None} & \multirow{2}{*}{-} & REMS & 21.48 & 13.90 & 20.21 & 22.75 & 26.08 & 22.30 & 19.03 \\
& & & EMS & 20.24 & 13.02 & 19.70 & 21.87 & 24.84 & 20.51 & 17.40 \\
\cmidrule{2-11}

& \multirow{6}{*}{Table} & \multirow{2}{*}{zero shots} & REMS & 49.59 & 33.52 & 40.18 & 63.60 & 62.07 & 47.53 & 43.62 \\
& & & EMS & 47.23 & 31.04 & 37.73 & 60.47 & 59.19 & 44.58 & 39.55 \\
\cmidrule{3-11}

& & \multirow{2}{*}{Static} & REMS & 49.94 & 36.94 & 41.33 & 66.45 & 64.23 & 51.13 & 47.13 \\
& & & EMS & 47.58 & 34.91 & 39.03 & 63.63 & 61.84 & 48.32 & 43.36 \\
\cmidrule{3-11}

& & \multirow{2}{*}{Adaptive} & REMS & 51.13 & 38.37 & 42.43 & 67.66 & 66.10 & 54.81 & 48.63 \\
& & & EMS & 48.79 & 35.48 & 39.96 & 64.80 & 63.16 & 52.07 & 44.49 \\
\midrule

\multirow{6}{*}{SQL} & \multirow{6}{*}{Schema} & \multirow{2}{*}{zero shots} & REMS & 39.93 & 41.28 & 38.76 & 48.50 & 57.45 & 50.88 & 34.21 \\
& & & EMS & 38.24 & 39.63 & 37.36 & 46.67 & 55.30 & 49.51 & 31.41 \\
\cmidrule{3-11}

& & \multirow{2}{*}{Static} & REMS & 57.44 & 51.18 & 53.94 & 66.52 & 77.57 & 65.29 & 57.04 \\
& & & EMS & 56.57 & 50.36 & 53.16 & 65.73 & 75.93 & 65.29 & 56.33 \\
\cmidrule{3-11}

& & \multirow{2}{*}{Adaptive} & REMS & 68.97 & 65.40 & 65.70 & 69.20 & 76.98 & 71.07 & 63.93 \\
& & & EMS & 68.69 & 65.24 & 65.43 & 68.77 & 76.87 & 70.41 & 63.13 \\
\bottomrule

\end{tabular}
\vspace{-1.0em}
\caption{\small Test Set Results for GPT-4o-mini}
\label{tab:gpt_4o_mini_results}
\vspace{-1.0em}
\end{table*}

\renewcommand{\arraystretch}{1.0}
\begin{table*}[h]
\centering
\tiny
\begin{tabular}{ll|c|c|ccccccccc}
\toprule
\multirow{2}{*}{\textbf{Output}} & \multirow{2}{*}{\textbf{Context}} & \multirow{2}{*}{\textbf{Method}} & \textbf{Metric} & \multicolumn{9}{c}{\textbf{Results Across Categories}} \\
\cmidrule{5-13}
& & & & \shortstack{\textbf{Original}} & \shortstack{\textbf{CounterFact}} & \shortstack{\textbf{Gap (Original} \\ \textbf{- CounterFact)}} & \shortstack{\textbf{Large}} & \shortstack{\textbf{Small}} & \shortstack{\textbf{Gap (Small} \\ \textbf{- Large)}} & \shortstack{\textbf{Easy}} & \shortstack{\textbf{Medium}} & \shortstack{\textbf{Hard}} \\
\midrule

\multirow{27}{*}{Text} & \multirow{2}{*}{None} & \multirow{2}{*}{-} & REMS & 22.11 & 15.02 & 7.09 & 23.88 & 21.74 & -2.14 & 27.81 & 23.23 & 21.59 \\
& & & EMS & 18.69 & 11.76 & 6.93 & 19.49 & 18.93 & -0.56 & 24.03 & 19.41 & 17.30 \\
\cmidrule{2-13}

& \multirow{22}{*}{Table} & \multirow{2}{*}{zero shots} & REMS & 55.17 & 45.42 & 9.75 & 42.71 & 69.88 & 27.17 & 68.78 & 56.61 & 51.18 \\
& & & EMS & 48.79 & 37.61 & 11.18 & 34.96 & 61.11 & 26.15 & 59.15 & 48.82 & 39.00 \\
\cmidrule{3-13}

& & \multirow{2}{*}{Static} & REMS & 57.46 & 47.20 & 10.26 & 46.44 & 71.89 & 25.45 & 73.76 & 61.94 & 54.99 \\
& & & EMS & 50.00 & 39.08 & 10.92 & 38.35 & 62.35 & 24.00 & 63.16 & 53.53 & 42.96 \\
\cmidrule{3-13}

& & \multirow{2}{*}{Adaptive} & REMS & 59.09 & 48.25 & 10.84 & 47.93 & 73.17 & 25.24 & 76.28 & 64.94 & 57.83 \\
& & & EMS & 52.90 & 42.91 & 9.99 & 41.02 & 66.25 & 25.23 & 65.76 & 55.95 & 45.92 \\
\cmidrule{3-13}

& & \multirow{2}{*}{Clear} & REMS & 62.01 & 50.46 & 11.55 & 53.38 & 73.07 & 19.69 & 79.65 & 68.02 & 59.37 \\
& & & EMS & 60.63 & 48.95 & 11.68 & 51.11 & 72.19 & 21.08 & 77.92 & 66.52 & 56.67 \\
\cmidrule{3-13}

& & \multirow{2}{*}{Chain Of Thought} & REMS & 60.97 & 47.28 & 13.69 & 51.61 & 73.45 & 21.84 & 77.28 & 65.60 & 56.62 \\
& & & EMS & 59.54 & 45.48 & 14.06 & 49.52 & 72.09 & 22.57 & 75.29 & 64.17 & 54.55 \\
\cmidrule{3-13}

& & \multirow{2}{*}{Plan And Solve} & REMS & 53.58 & 42.58 & 11.00 & 48.40 & 65.07 & 16.67 & 70.61 & 59.35 & 51.43 \\
& & & EMS & 52.00 & 40.34 & 11.66 & 46.14 & 63.85 & 17.71 & 68.54 & 57.12 & 48.53 \\
\cmidrule{3-13}

& & \multirow{2}{*}{Program of Thought} & REMS & 49.26 & 42.85 & 6.41 & 44.31 & 57.29 & 12.98 & 67.10 & 48.61 & 44.13 \\
& & & EMS & 46.90 & 40.55 & 6.35 & 42.33 & 55.30 & 12.97 & 64.53 & 45.52 & 40.47 \\
\cmidrule{3-13}

& & \multirow{2}{*}{Faithful CoT} & REMS & 48.98 & 40.90 & 8.08 & 43.46 & 58.09 & 14.63 & 64.12 & 53.26 & 45.68 \\
& & & EMS & 46.78 & 38.66 & 8.12 & 41.38 & 56.23 & 14.85 & 61.50 & 50.66 & 42.67 \\

\midrule

\multirow{6}{*}{SQL} & \multirow{6}{*}{Schema} & \multirow{2}{*}{zero shots} & REMS & 47.27 & 42.56 & 4.71 & 46.23 & 56.45 & 10.22 & 63.49 & 59.74 & 42.09 \\
& & & EMS & 39.34 & 33.45 & 5.89 & 44.29 & 53.65 & 9.36 & 52.96 & 52.48 & 34.59 \\
\cmidrule{3-13}

& & \multirow{2}{*}{Static} & REMS & 66.43 & 62.38 & 4.05 & 71.39 & 77.20 & 5.81 & 88.17 & 79.22 & 65.04 \\
& & & EMS & 57.93 & 54.43 & 3.50 & 63.13 & 70.89 & 7.76 & 79.54 & 68.96 & 58.42 \\
\cmidrule{3-13}

& & \multirow{2}{*}{Adaptive} & REMS & 72.91 & 71.67 & 1.24 & 78.18 & 81.71 & 3.53 & 87.06 & 80.80 & 74.04 \\
& & & EMS & 65.49 & 62.71 & 2.78 & 69.30 & 73.53 & 4.23 & 76.26 & 73.72 & 63.12 \\
\bottomrule

\end{tabular}
\caption{\small Test Set Results for Gemini 1.5 Flash.}
\vspace{-1.0em}
\label{tab:gemini_1_5_flash_results}
\vspace{-1.0em}
\end{table*}

\renewcommand{\arraystretch}{1.0}
\begin{table*}[h]
\centering
\tiny
\begin{tabular}{ll|c|c|ccccccccc}
\toprule
\multirow{1}{*}{\textbf{Output}} & \multirow{1}{*}{\textbf{Context}} & \multirow{1}{*}{\textbf{Method}} & \textbf{Metric} & \textbf{Original} & \textbf{CounterFact} & \shortstack{\textbf{Gap (Original} \\ \textbf{- CounterFact)}} & \textbf{Large} & \textbf{Small} & \shortstack{\textbf{Gap (Small} \\ \textbf{- Large)}} & \textbf{Easy} & \textbf{Medium} & \textbf{Hard} \\
\midrule

\multirow{27}{*}{Text} 

\multirow{2}{*}{} & \multirow{2}{*}{None} & \multirow{2}{*}{-} & REMS & 23.37 & 15.88 & 7.49 & 24.20 & 24.08 & 0.12 & 29.15 & 26.45 & 23.38 \\
& & & EMS & 19.90 & 13.45 & 6.45 & 21.19 & 20.99 & 0.2 & 24.94 & 21.76 & 18.77 \\
\cmidrule{2-13}

& \multirow{6}{*}{} & \multirow{2}{*}{zero shots} & REMS & 54.94 & 45.40 & 9.54 & 46.34 & 69.74 & 23.4 & 73.26 & 60.71 & 57.43 \\
& & & EMS & 48.06 & 37.39 & 10.67 & 38.98 & 61.11 & 21.13 & 62.59 & 53.24 & 45.60 \\
\cmidrule{3-13}

& & \multirow{2}{*}{Static} & REMS & 59.01 & 46.87 & 12.14 & 52.42 & 72.93 & 20.51 & 75.72 & 65.10 & 60.28 \\
& & & EMS & 52.91 & 39.92 & 12.99 & 43.86 & 65.02 & 21.16 & 65.79 & 58.53 & 50.00 \\
\cmidrule{3-13}

& & \multirow{2}{*}{Adaptive} & REMS & 60.23 & 49.04 & 11.19 & 48.79 & 73.98 & 25.19 & 78.04 & 66.43 & 59.28 \\
& & & EMS & 53.48 & 44.19 & 9.29 & 41.86 & 67.27 & 25.41 & 66.26 & 56.47 & 46.74 \\
\cmidrule{3-13}

& \multirow{5}{*}{Table} & \multirow{2}{*}{Clear} & REMS & 50.29 & 42.82 & 7.47 & 42.83 & 58.04 & 15.21 & 59.91 & 60.79 & 52.82 \\
& & & EMS & 49.33 & 40.66 & 8.67 & 40.95 & 56.85 & 15.90 & 58.38 & 59.77 & 51.14 \\
\cmidrule{3-13}

& & \multirow{2}{*}{Chain Of Thought} & REMS & 67.34 & 57.16 & 10.18 & 59.51 & 78.47 & 18.96 & 84.67 & 73.36 & 67.31 \\
& & & EMS & 66.46 & 55.75 & 10.71 & 57.67 & 77.55 & 19.88 & 83.29 & 72.69 & 65.87 \\
\cmidrule{3-13}

& & \multirow{2}{*}{Plan And Solve} & REMS & 62.11 & 55.68 & 6.43 & 58.38 & 74.88 & 16.50 & 81.19 & 73.28 & 65.56 \\
& & & EMS & 60.75 & 54.31 & 6.44 & 56.72 & 73.43 & 16.71 & 79.42 & 72.25 & 63.60 \\
\cmidrule{3-13}

& & \multirow{2}{*}{Program of Thought} & REMS & 56.31 & 49.35 & 6.97 & 53.02 & 66.46 & 13.44 & 76.94 & 60.62 & 51.75 \\
& & & EMS & 53.83 & 46.20 & 7.63 & 49.84 & 64.68 & 14.83 & 73.97 & 57.12 & 47.57 \\
\cmidrule{3-13}

& & \multirow{2}{*}{Faithful CoT} & REMS & 56.49 & 50.15 & 6.34 & 50.63 & 65.84 & 15.21 & 77.28 & 61.53 & 51.03 \\
& & & EMS & 53.95 & 47.64 & 6.31 & 48.47 & 63.85 & 15.39 & 74.52 & 58.15 & 46.85 \\

\midrule

\multirow{6}{*}{SQL} 
& \multirow{6}{*}{Schema} & \multirow{2}{*}{zero shots} & REMS & 49.24 & 43.56 & 5.68 & 48.21 & 57.62 & 9.41 & 64.42 & 61.10 & 43.00 \\
& & & EMS & 41.32 & 35.31 & 6.01 & 45.87 & 55.11 & 9.24 & 54.78 & 53.79 & 35.96 \\
\cmidrule{3-13}

& & \multirow{2}{*}{Static} & REMS & 67.76 & 63.63 & 4.13 & 76.80 & 82.82 & 6.02 & 89.33 & 80.81 & 66.42 \\
& & & EMS & 59.08 & 55.52 & 3.56 & 71.32 & 77.41 & 6.09 & 80.86 & 70.33 & 59.59 \\
\cmidrule{3-13}

& & \multirow{2}{*}{Adaptive} & REMS & 73.04 & 73.58 & 0.54 & 81.94 & 84.38 & 2.44 & 87.31 & 80.01 & 71.43 \\
& & & EMS & 65.29 & 65.13 & 0.16 & 72.43 & 75.31 & 2.88 & 75.86 & 71.47 & 59.24 \\

\bottomrule
\end{tabular}
\vspace{-1.0em}
\caption{\small Test Set Results for Gemini 1.5 Pro.}
\label{tab:gemini_1_5_pro_results}
\vspace{-1.0em}
\end{table*}

\begin{table*}[h]
\centering
\tiny
\begin{tabular}{ll|c|c|ccccccccc}
\toprule
\multirow{2}{*}{\textbf{Output}} & \multirow{2}{*}{\textbf{Context}} & \multirow{2}{*}{\textbf{Method}} & \textbf{Metric} & \multicolumn{9}{c}{\textbf{Results Across Categories}} \\
\cmidrule{5-13}
& & & & \shortstack{\textbf{Original}} & \shortstack{\textbf{CounterFact}} & \shortstack{\textbf{Gap (Original} \\ \textbf{- CounterFact)}} & \shortstack{\textbf{Large}} & \shortstack{\textbf{Small}} & \shortstack{\textbf{Gap (Small} \\ \textbf{- Large)}} & \shortstack{\textbf{Easy}} & \shortstack{\textbf{Medium}} & \shortstack{\textbf{Hard}} \\
\midrule

\multirow{27}{*}{Text} & \multirow{2}{*}{None} & \multirow{2}{*}{-} & REMS & 17.46 & 12.74 & 4.72 & 20.89 & 18.09 & -2.80 & 22.31 & 17.93 & 17.18 \\
& & & EMS & 16.61 & 12.16 & 4.45 & 19.89 & 17.31 & -2.58 & 21.31 & 16.96 & 16.27 \\
\cmidrule{2-13}

& \multirow{22}{*}{Table} & \multirow{2}{*}{zero shots} & REMS & 54.63 & 39.93 & 14.70 & 46.10 & 64.62 & 18.52 & 70.10 & 58.31 & 53.37 \\
& & & EMS & 52.60 & 37.48 & 15.12 & 44.05 & 62.57 & 18.52 & 67.49 & 56.21 & 50.35 \\
\cmidrule{3-13}

& & \multirow{2}{*}{Static} & REMS & 64.33 & 48.44 & 15.89 & 57.27 & 75.40 & 18.13 & 79.04 & 70.23 & 60.98 \\
& & & EMS & 62.46 & 46.21 & 16.25 & 55.58 & 73.68 & 18.10 & 76.91 & 67.46 & 57.44 \\
\cmidrule{3-13}

& & \multirow{2}{*}{Adaptive} & REMS & 55.73 & 41.29 & 14.44 & 48.48 & 70.56 & 22.08 & 71.35 & 62.54 & 56.43 \\
& & & EMS & 53.63 & 39.20 & 14.43 & 46.65 & 68.07 & 21.42 & 69.40 & 59.76 & 52.85 \\
\cmidrule{3-13}

& & \multirow{2}{*}{Clear} & REMS & 54.37 & 39.43 & 14.94 & 37.63 & 68.64 & 31.01 & 68.86 & 62.91 & 50.87 \\
& & & EMS & 53.46 & 38.34 & 15.12 & 36.43 & 67.02 & 30.59 & 67.49 & 62.33 & 49.51 \\
\cmidrule{3-13}

& & \multirow{2}{*}{Chain Of Thought} & REMS & 61.87 & 45.29 & 16.58 & 49.87 & 73.88 & 24.01 & 77.68 & 70.40 & 59.48 \\
& & & EMS & 59.86 & 43.06 & 16.80 & 48.51 & 72.04 & 23.53 & 75.13 & 69.23 & 57.16 \\
\cmidrule{3-13}

& & \multirow{2}{*}{Plan And Solve} & REMS & 59.31 & 45.13 & 14.18 & 50.55 & 71.82 & 21.27 & 75.65 & 67.06 & 57.32 \\
& & & EMS & 57.79 & 43.49 & 14.30 & 49.07 & 70.29 & 21.22 & 73.63 & 66.08 & 55.21 \\
\cmidrule{3-13}

& & \multirow{2}{*}{Program of Thought} & REMS & 46.63 & 31.83 & 14.80 & 28.30 & 63.03 & 34.73 & 61.16 & 51.98 & 41.83 \\
& & & EMS & 44.64 & 29.90 & 14.74 & 27.69 & 61.40 & 33.71 & 59.15 & 48.91 & 39.36 \\
\cmidrule{3-13}

& & \multirow{2}{*}{Faithful CoT} & REMS & 40.19 & 29.44 & 10.75 & 25.11 & 60.30 & 35.19 & 56.52 & 46.22 & 37.86 \\
& & & EMS & 38.75 & 27.75 & 11.00 & 24.54 & 58.83 & 34.29 & 55.19 & 43.79 & 35.74 \\

\midrule

\multirow{6}{*}{SQL} & \multirow{6}{*}{Schema} & \multirow{2}{*}{zero shots} & REMS & 34.77 & 31.52 & 3.25 & 37.08 & 41.36 & 4.28 & 45.83 & 40.81 & 29.74 \\
& & & EMS & 33.56 & 30.47 & 3.09 & 36.06 & 39.77 & 3.71 & 44.26 & 39.65 & 27.54 \\
\cmidrule{3-13}

& & \multirow{2}{*}{Static} & REMS & 54.59 & 47.31 & 7.28 & 55.38 & 63.12 & 7.74 & 68.51 & 60.83 & 56.15 \\
& & & EMS & 53.81 & 46.64 & 7.17 & 54.65 & 62.22 & 7.57 & 67.21 & 60.55 & 55.63 \\
\cmidrule{3-13}

& & \multirow{2}{*}{Adaptive} & REMS & 64.61 & 64.04 & 0.57 & 67.40 & 69.54 & 2.14 & 77.42 & 68.01 & 61.78 \\
& & & EMS & 64.36 & 63.81 & 0.55 & 66.91 & 68.89 & 1.98 & 77.19 & 67.65 & 60.92 \\
\bottomrule

\end{tabular}
\vspace{-1.0em}
\caption{\small Test Set Results for Llama 3.1 70B.}
\label{tab:llama_3_1_70b_results}
\vspace{-1.0em}
\end{table*}

\begin{table*}[h]
\centering
\small
\begin{tabular}{ll|c|c|ccccccc}
\toprule
\multirow{2}{*}{\textbf{Output}} & \multirow{2}{*}{\textbf{Context}} & \multirow{2}{*}{\textbf{Few Shots}} & \textbf{Metric} & \multicolumn{7}{c}{\textbf{Results Across Categories}} \\
\cmidrule{5-11}
& & & & \shortstack{\textbf{Original}} & \shortstack{\textbf{CounterFact}} & \shortstack{\textbf{Large}} & \shortstack{\textbf{Small}} & \shortstack{\textbf{Easy}} & \shortstack{\textbf{Medium}} & \shortstack{\textbf{Hard}} \\
\midrule

\multirow{11}{*}{Text} & \multirow{2}{*}{None} & \multirow{2}{*}{-} & REMS & 17.53 & 13.52 & 21.71 & 20.31 & 20.38 & 20.06 & 19.95 \\
& & & EMS & 15.92 & 12.73 & 20.82 & 19.53 & 19.26 & 17.75 & 18.08 \\
\cmidrule{2-11}

& \multirow{6}{*}{Table} & \multirow{2}{*}{zero shots} & REMS & 40.26 & 32.62 & 37.58 & 56.55 & 54.09 & 46.27 & 38.65 \\
& & & EMS & 38.06 & 30.47 & 35.50 & 53.33 & 51.78 & 43.79 & 35.05 \\
\cmidrule{3-11}

& & \multirow{2}{*}{Static} & REMS & 50.92 & 42.31 & 43.17 & 68.13 & 64.75 & 55.80 & 49.06 \\
& & & EMS & 48.79 & 40.20 & 40.89 & 65.03 & 62.30 & 53.25 & 45.62 \\
\cmidrule{3-11}

& & \multirow{2}{*}{Adaptive} & REMS & 39.98 & 33.23 & 37.68 & 50.69 & 53.84 & 41.02 & 40.33 \\
& & & EMS & 37.54 & 30.62 & 34.94 & 47.72 & 50.96 & 38.46 & 35.74 \\
\midrule

\multirow{6}{*}{SQL} & \multirow{6}{*}{Schema} & \multirow{2}{*}{zero shots} & REMS & 20.35 & 15.07 & 19.30 & 27.74 & 21.65 & 27.19 & 22.43 \\
& & & EMS & 19.90 & 14.74 & 18.96 & 26.55 & 20.63 & 27.02 & 21.56 \\
\cmidrule{3-11}

& & \multirow{2}{*}{Static} & REMS & 47.42 & 41.38 & 48.78 & 54.63 & 69.46 & 46.22 & 42.45 \\
& & & EMS & 46.02 & 40.20 & 47.96 & 52.98 & 67.35 & 45.36 & 40.61 \\
\cmidrule{3-11}

& & \multirow{2}{*}{Adaptive} & REMS & 25.45 & 21.26 & 24.72 & 33.96 & 26.91 & 47.24 & 22.20 \\
& & & EMS & 25.09 & 20.89 & 24.54 & 33.57 & 26.78 & 46.55 & 21.56 \\
\bottomrule

\end{tabular}
\vspace{-1.0em}
\caption{\small Test Set Results for Mixtral 8x7B.}
\vspace{-1.0em}
\label{tab:mixtral_8x7b_results}
\end{table*}

\begin{table*}[h]
\centering
\footnotesize
\begin{tabular}{ll|c|c|ccccccc}
\toprule
\multirow{2}{*}{\textbf{Output}} & \multirow{2}{*}{\textbf{Context}} & \multirow{2}{*}{\textbf{Few Shots}} & \textbf{Metric} & \multicolumn{7}{c}{\textbf{Results Across Categories}} \\
\cmidrule{5-11}
& & & & \shortstack{\textbf{Original}} & \shortstack{\textbf{CounterFact}} & \shortstack{\textbf{Large}} & \shortstack{\textbf{Small}} & \shortstack{\textbf{Easy}} & \shortstack{\textbf{Medium}} & \shortstack{\textbf{Hard}} \\
\midrule

\multirow{6}{*}{SQL} & \multirow{6}{*}{Schema} & \multirow{2}{*}{zero shots} & REMS & 19.59 & 18.32 & 20.56 & 27.60 & 29.05 & 20.28 & 17.29 \\
& & & EMS & 18.17 & 17.02 & 19.46 & 26.08 & 26.78 & 19.72 & 15.72 \\
\cmidrule{3-11}

& & \multirow{2}{*}{Static} & REMS & 53.37 & 49.32 & 63.43 & 61.01 & 77.39 & 54.05 & 30.19 \\
& & & EMS & 52.08 & 47.93 & 62.28 & 59.53 & 77.39 & 51.19 & 28.91 \\
\cmidrule{3-11}

& & \multirow{2}{*}{Adaptive} & REMS & 57.10 & 55.03 & 64.00 & 60.55 & 65.02 & 60.13 & 54.07 \\
& & & EMS & 55.88 & 53.50 & 63.17 & 58.95 & 63.93 & 58.38 & 51.74 \\
\bottomrule

\end{tabular}
\vspace{-1.0em}
\caption{\small Test Set Results for SQL Coder 70B}
\label{tab:sql_coder_results}
\end{table*}

\begin{table*}[h]
\centering
\footnotesize
\begin{tabular}{ll|c|c|ccccccc}
\toprule
\multirow{2}{*}{\textbf{Output}} & \multirow{2}{*}{\textbf{Context}} & \multirow{2}{*}{\textbf{Few Shots}} & \textbf{Metric} & \multicolumn{7}{c}{\textbf{Results Across Categories}} \\
\cmidrule{5-11}
& & & & \shortstack{\textbf{Original}} & \shortstack{\textbf{CounterFact}} & \shortstack{\textbf{Large}} & \shortstack{\textbf{Small}} & \shortstack{\textbf{Easy}} & \shortstack{\textbf{Medium}} & \shortstack{\textbf{Hard}} \\
\midrule

\multirow{6}{*}{SQL} & \multirow{6}{*}{Schema} & \multirow{2}{*}{zero shot} & REMS & 12.37 & 19.77 & 13.76 & 21.31 & 33.81 & 25.98 & 19.19 \\
& & & EMS & 12.22 & 19.46 & 13.76 & 21.00 & 33.47 & 25.64 & 18.64 \\
\cmidrule{3-11}

& & \multirow{2}{*}{Static} & REMS & 16.53 & 33.29 & 29.82 & 42.04 & 54.13 & 41.62 & 39.09 \\
& & & EMS & 15.84 & 32.62 & 29.82 & 41.64 & 53.42 & 41.62 & 38.94 \\
\cmidrule{3-11}

& & \multirow{2}{*}{Adaptive} & REMS & 24.02 & 40.73 & 38.00 & 48.56 & 65.56 & 51.78 & 41.63 \\
& & & EMS & 23.53 & 40.06 & 37.61 & 48.14 & 65.16 & 50.89 & 40.61 \\
\bottomrule

\end{tabular}
\vspace{-1.0em}
\caption{\small Test Set Results for Code Llama.}
\label{tab:code_llama_results}
\vspace{-1.0em}
\end{table*}

\onecolumn
\section{SQL Code Generation Prompt}

\noindent \textbf{\# Task Instruction:} \\
You will be given a question and your task is to provide the SQL logic to answer a natural language question based on the provided schema. Few Examples of the task will be provided below. Assume that all the data is already inserted into the database.

\noindent
\textbf{\small 1. Table Schemas:}
\footnotesize
\begin{lstlisting}[language=SQL]
CREATE TABLE Athlete (
    athlete_id INT AUTO_INCREMENT PRIMARY KEY,
    name VARCHAR(100) NOT NULL
);
CREATE TABLE Tournament (
    tournament_id INT AUTO_INCREMENT PRIMARY KEY,
    athlete_id INT,
    name VARCHAR(100) NOT NULL,
    FOREIGN KEY (athlete_id) REFERENCES Athlete(athlete_id)
);
CREATE TABLE Format (
    format_id INT AUTO_INCREMENT PRIMARY KEY,
    tournament_id INT,
    name VARCHAR(100) NOT NULL,
    FOREIGN KEY (tournament_id) REFERENCES Tournament(tournament_id)
);
CREATE TABLE Medal (
    medal_id INT AUTO_INCREMENT PRIMARY KEY,
    format_id INT,
    type VARCHAR(50) NOT NULL,
    year INT,
    location VARCHAR(100) NOT NULL,
    FOREIGN KEY (format_id) REFERENCES Format(format_id)
);
CREATE TABLE PersonalInformation (
    info_id INT AUTO_INCREMENT PRIMARY KEY,
    athlete_id INT,
    birth_year INT,
    birth_month INT,
    birth_day INT,
    FOREIGN KEY (athlete_id) REFERENCES Athlete(athlete_id)
);
\end{lstlisting}

\noindent
\textbf{2. Table Descriptions:}
\begin{verbatim}
describe athlete;
+------------+--------------+------+-----+---------+----------------+
| Field      | Type         | Null | Key | Default | Extra          |
+------------+--------------+------+-----+---------+----------------+
| athlete_id | int(11)      | NO   | PRI | NULL    | auto_increment |
| name       | varchar(100) | NO   |     | NULL    |                |
+------------+--------------+------+-----+---------+----------------+
describe personalinformation;
+-------------+---------+------+-----+---------+----------------+
| Field       | Type    | Null | Key | Default | Extra          |
+-------------+---------+------+-----+---------+----------------+
| info_id     | int(11) | NO   | PRI | NULL    | auto_increment |
| athlete_id  | int(11) | YES  | MUL | NULL    |                |
| birth_year  | int(11) | YES  |     | NULL    |                |
| birth_month | int(11) | YES  |     | NULL    |                |
| birth_day   | int(11) | YES  |     | NULL    |                |
+-------------+---------+------+-----+---------+----------------+
describe tournament;
+---------------+--------------+------+-----+---------+----------------+
| Field         | Type         | Null | Key | Default | Extra          |
+---------------+--------------+------+-----+---------+----------------+
| tournament_id | int(11)      | NO   | PRI | NULL    | auto_increment |
| athlete_id    | int(11)      | YES  | MUL | NULL    |                |
| name          | varchar(100) | NO   |     | NULL    |                |
+---------------+--------------+------+-----+---------+----------------+
describe format;
+---------------+--------------+------+-----+---------+----------------+
| Field         | Type         | Null | Key | Default | Extra          |
+---------------+--------------+------+-----+---------+----------------+
| format_id     | int(11)      | NO   | PRI | NULL    | auto_increment |
| tournament_id | int(11)      | YES  | MUL | NULL    |                |
| name          | varchar(100) | NO   |     | NULL    |                |
+---------------+--------------+------+-----+---------+----------------+
describe medal;
+-----------+--------------+------+-----+---------+----------------+
| Field     | Type         | Null | Key | Default | Extra          |
+-----------+--------------+------+-----+---------+----------------+
| medal_id  | int(11)      | NO   | PRI | NULL    | auto_increment |
| format_id | int(11)      | YES  | MUL | NULL    |                |
| type      | varchar(50)  | NO   |     | NULL    |                |
| year      | int(11)      | YES  |     | NULL    |                |
| location  | varchar(100) | NO   |     | NULL    |                |
+-----------+--------------+------+-----+---------+----------------+
\end{verbatim}

\vspace{10pt}
\noindent
\textbf{3. Example Data:}
\begin{verbatim}
Athlete Table
+------------+-----------------+
| athlete_id | name            |
+------------+-----------------+
|         50 | Carolina Marín  |
+------------+-----------------+
PersonalInformation Table
+---------+------------+------------+-------------+-----------+
| info_id | athlete_id | birth_year | birth_month | birth_day |
+---------+------------+------------+-------------+-----------+
|      40 |         50 |       1993 |           6 |        15 |
+---------+------------+------------+-------------+-----------+
Tournament Table
+---------------+------------+-------------------------------+
| tournament_id | athlete_id | name                          |
+---------------+------------+-------------------------------+
|           281 |         50 | Olympic Games                 |
|           282 |         50 | World Championships           |
|           285 |         50 | European Women                |
+---------------+------------+-------------------------------+
Format Table
+-----------+---------------+------------------+
| format_id | tournament_id | name             |
+-----------+---------------+------------------+
|       392 |           281 | Women's singles  |
|       393 |           282 | Women's singles  |
|       396 |           285 | Women's team     |
+-----------+---------------+------------------+
Medal Table
+----------+-----------+-------------+------+----------------+
| medal_id | format_id | type        | year | location       |
+----------+-----------+-------------+------+----------------+
|      692 |       392 | MedalGold   | 2016 | Rio de Janeiro |
|      696 |       393 | MedalSilver | 2023 | Copenhagen     |
|      706 |       396 | MedalBronze | 2016 | Kazan          |
+----------+-----------+-------------+------+----------------+
\end{verbatim}

\vspace{10pt}
\noindent
\textbf{Example 1:} \\
\textit{Question: Which tournament(s) has Zhang Jike won the most Medals in?}
\begin{lstlisting}[language=SQL]
WITH medal_counts AS (
    SELECT t.name AS tournament_name, m.year, COUNT(m.medal_id) AS total_medal_count
    FROM Medal m
    JOIN Format f ON m.format_id = f.format_id
    JOIN Tournament t ON f.tournament_id = t.tournament_id
    JOIN Athlete a ON t.athlete_id = a.athlete_id
    WHERE a.name = 'Zhang Jike'
    GROUP BY t.name, m.year
)
SELECT tournament_name, year
FROM medal_counts
WHERE total_medal_count = (
    SELECT MAX(total_medal_count)
    FROM medal_counts
);
\end{lstlisting}

\vspace{10pt}
\noindent
\textbf{Example 2:} \\
\textit{Question: In which year(s) did Seo Seung-jae win medals in the Asian Junior Championships?}
\begin{lstlisting}[language=SQL]
SELECT DISTINCT m.year
FROM Medal m
JOIN Format f ON m.format_id = f.format_id
JOIN Tournament t ON f.tournament_id = t.tournament_id
JOIN Athlete a ON t.athlete_id = a.athlete_id
WHERE a.name = 'Seo Seung-jae'
  AND t.name = 'Asian Junior Championships';
\end{lstlisting}

\vspace{10pt}
\noindent
\textbf{Example 3:} \\
\textit{Question: Which was the most current medal win for Dola Banerjee?}
\begin{lstlisting}[language=SQL]
SELECT m.type, m.year, m.location, f.name AS format_name, t.name AS tournament_name
FROM Medal m
JOIN Format f ON m.format_id = f.format_id
JOIN Tournament t ON f.tournament_id = t.tournament_id
JOIN Athlete a ON t.athlete_id = a.athlete_id
WHERE a.name = 'Dola Banerjee'
  AND m.year = (
    SELECT MAX(m2.year)
    FROM Medal m2
    JOIN Format f2 ON m2.format_id = f2.format_id
    JOIN Tournament t2 ON f2.tournament_id = t2.tournament_id
    JOIN Athlete a2 ON t2.athlete_id = a2.athlete_id
    WHERE a2.name = 'Dola Banerjee'
  );
\end{lstlisting}

\vspace{10pt}
\noindent
\textbf{Example 4:} \\
\textit{Question: How many international medals did Rawinda Prajongjai win in 2023?}
\begin{lstlisting}[language=SQL]
SELECT COUNT(m.medal_id) AS total_medals
FROM Medal m
JOIN Format f ON m.format_id = f.format_id
JOIN Tournament t ON f.tournament_id = t.tournament_id
JOIN Athlete a ON t.athlete_id = a.athlete_id
WHERE a.name = 'Rawinda Prajongjai'
  AND m.year = 2023;
\end{lstlisting}

\vspace{10pt}
\noindent
\textbf{Example 5:} \\
\textit{Question: In which year(s) did Huang Dongping win the highest number of medals during their career?}
\begin{lstlisting}[language=SQL]
SELECT m.year
FROM Medal m
JOIN Format f ON m.format_id = f.format_id
JOIN Tournament t ON f.tournament_id = t.tournament_id
JOIN Athlete a ON t.athlete_id = a.athlete_id
WHERE a.name = 'Huang Dongping'
GROUP BY m.year
ORDER BY COUNT(m.medal_id) DESC
LIMIT 1;
\end{lstlisting}

\vspace{10pt}
\noindent
\textbf{Example 6:} \\
\textit{Question: In which year(s) did Tomokazu Harimoto win the lowest number of medals during their career?}
\begin{lstlisting}[language=SQL]
SELECT m.year
FROM Medal m
JOIN Format f ON m.format_id = f.format_id
JOIN Tournament t ON f.tournament_id = t.tournament_id
JOIN Athlete a ON t.athlete_id = a.athlete_id
WHERE a.name = 'Tomokazu Harimoto'
GROUP BY m.year
HAVING COUNT(m.medal_id) = (
    SELECT MIN(medal_count)
    FROM (
        SELECT COUNT(m2.medal_id) AS medal_count
        FROM Medal m2
        JOIN Format f2 ON m2.format_id = f2.format_id
        JOIN Tournament t2 ON f2.tournament_id = t2.tournament_id
        JOIN Athlete a2 ON t2.athlete_id = a2.athlete_id
        WHERE a2.name = 'Tomokazu Harimoto'
        GROUP BY m2.year
    ) AS yearly_medal_counts
);
\end{lstlisting}

\vspace{10pt}
\noindent
\textbf{Instructions for Writing Queries:}
\begin{enumerate}
    \item If a question can have multiple answers, do not limit the response to only one. Instead, output all possible answers.
    \item Use the column names as specified in the schema to find the necessary parameters for the query.
    \item An event is a combination of Tournament, Format, and the corresponding year.
    \item There are three types of medals in the Medal Table: MedalGold, MedalSilver, MedalBronze.
\end{enumerate}

\end{document}